\documentclass[sigconf]{acmart}

\AtBeginDocument{%
  }

\copyrightyear{2025}
\acmYear{2025}
\setcopyright{acmlicensed}
\acmConference[MM '25] {Proceedings of the 33rd ACM International Conference on Multimedia}{October 27--31, 2025}{Dublin, Ireland.}
\acmBooktitle{Proceedings of the 33rd ACM International Conference on Multimedia (MM '25), October 27--31, 2025, Dublin, Ireland}
\acmISBN{979-8-4007-2035-2/2025/10}
\acmDOI{10.1145/3746027.3754912}

\acmSubmissionID{1104}

\usepackage{bibentry}
\usepackage{booktabs}
\usepackage{makecell}
\usepackage{multirow}
\usepackage{pifont}
\usepackage{tabularx}
\usepackage{graphicx}
\usepackage{amsmath}
\usepackage{threeparttable}

\usepackage{amsfonts,amssymb}
\usepackage[american]{babel}
\usepackage{colortbl, booktabs}
\usepackage{hhline}

\begin{document}

\title{Revealing Latent Information: A Physics-inspired Self-supervised Pre-training Framework for Noisy and Sparse Events}

\author{Lin Zhu}
\affiliation{%
  \institution{Beijing Institute of Technology}
  \city{Beijing}
  \country{China}}
\email{linzhu@bit.edu.cn}

\author{Ruonan Liu}
\affiliation{%
  \institution{Beijing Institute of Technology}
  \city{Beijing}
  \country{China}}
\email{liuruonan@bit.edu.cn}

\author{Xiao Wang}
\affiliation{%
 \institution{Anhui University}
 \city{Anhui}
 \country{China}}
\email{wangxiaocvpr@foxmail.com}

\author{Lizhi Wang}
\affiliation{%
  \institution{Beijing Normal University}
  \city{Beijing}
  \country{China}}
\email{wanglizhi@bnu.edu.cn}

\author{Hua Huang}\thanks{Corresponding author: Hua Huang}
\affiliation{%
  \institution{Beijing Normal University}
  \city{Beijing}
  \country{China}}
\email{huahuang@bnu.edu.cn}

\renewcommand{\shortauthors}{Zhu et al.}

\begin{abstract}
Event camera, a novel neuromorphic vision sensor, records data with high temporal resolution and wide dynamic range, offering new possibilities for accurate visual representation in challenging scenarios.
However, event data is inherently sparse and noisy, mainly reflecting brightness changes, which complicates effective feature extraction.
To address this, we propose a self-supervised pre-training framework to fully reveal latent information in event data, including edge information and texture cues.
Our framework consists of three stages:
Difference-guided Masked Modeling, inspired by the event physical sampling process, reconstructs temporal intensity difference maps to extract enhanced information from raw event data.
Backbone-fixed Feature Transition contrasts event and image features without updating the backbone to preserve representations learned from masked modeling and stabilizing their effect on contrastive learning.
Focus-aimed Contrastive Learning updates the entire model to improve semantic discrimination by focusing on high-value regions.
Extensive experiments show our framework is robust and consistently outperforms state-of-the-art methods on various downstream tasks, including object recognition, semantic segmentation, and optical flow estimation.
The code and dataset are available at \url{https://github.com/BIT-Vision/EventPretrain}.
\end{abstract}


\begin{teaserfigure}
  \vspace{-1mm}
  \includegraphics[width=\textwidth]{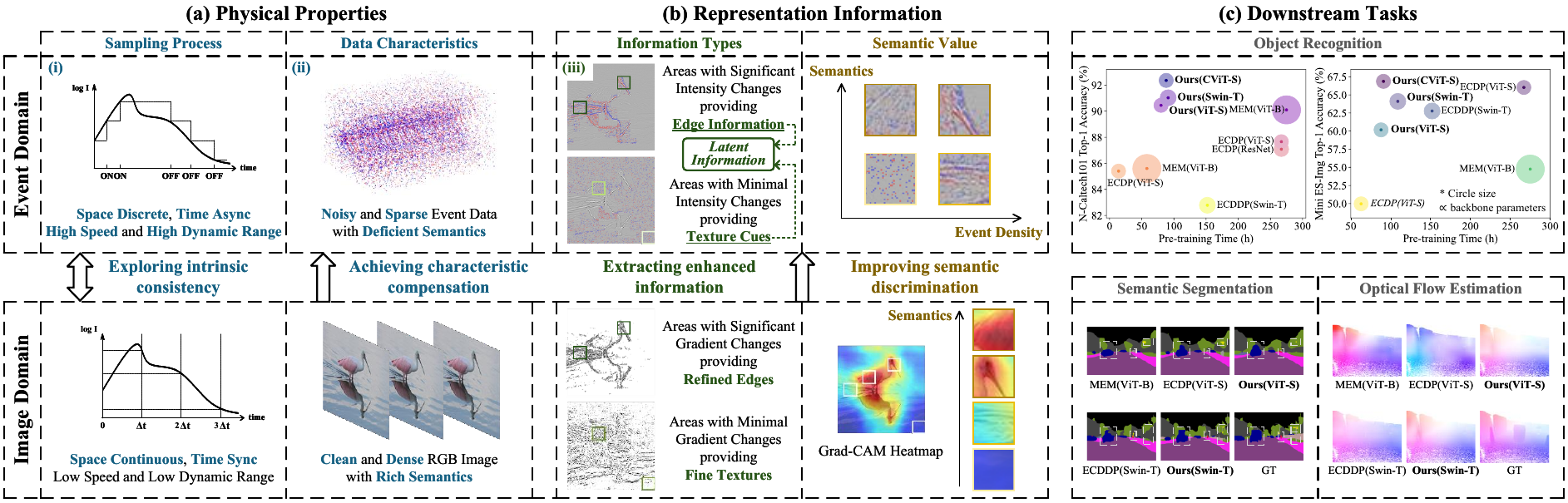}
  \vspace{-7mm}
  \caption{
  Learning high-quality event representations by revealing latent information through leveraging event-image domain commonalities.
  (a) Physical property comparison between events and images.
  (b) Representation information analysis of events with corresponding image.
  (c) Effectiveness of our method across various downstream tasks.
  }
  \label{introduction}
\end{teaserfigure}

\maketitle

\section{Introduction}
\label{sec:introduction}

\begin{figure*}[t]
\centering
\includegraphics[width=0.9\textwidth]{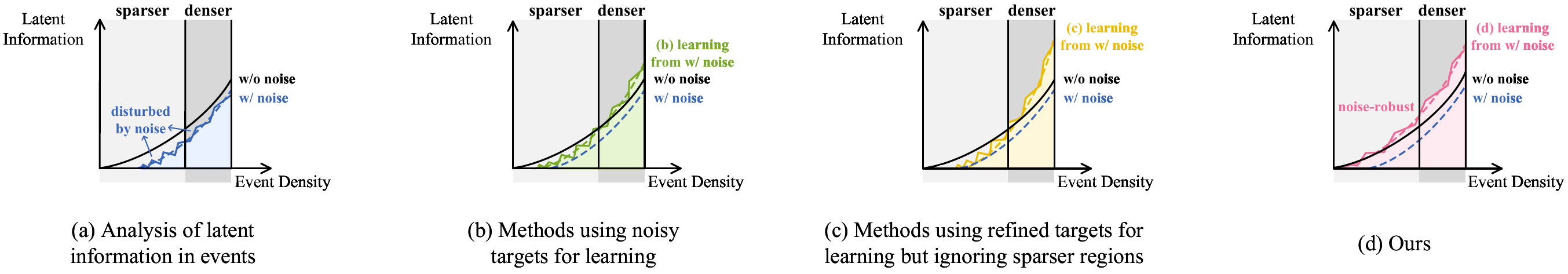}
\vspace{-4mm}
\caption{
The comparison of different event-based pre-training strategies.
(a) For ideal events without noise, denser areas cover a smaller range but convey more information than sparser areas. For real events with noise, information degrades due to noise interference, especially in sparser regions.
(b), (c) and (d): Through representation learning, the model can extract more potential information from the events' original limited information.
(b) Methods that use entire original events for learning, such as ECDDP~\cite{yang2024event} and MEM~\cite{klenk2024masked}, tend to broadly extract optimized information from all regions but are prone to information loss due to noise and sparsity.
(c) Methods that use image features or refined voxel features for learning, but mask or discard sparser areas, such as ECDP~\cite{yang2023event} and Huang et al.~\cite{huang2024data}, tend to extract more optimized information from denser regions but risk missing potential information in sparser areas.
(d) Our method sequentially learns from temporal intensity difference maps and image features, exhibiting robustness to noise and effectively extracting optimized information from both dense and sparse regions, thereby fully leveraging potential event information for enhanced representation.
}
\vspace{-2mm}
\label{compare}
\end{figure*}

The event camera~\cite{brandli2014240,delbruck2010activity}, a novel neuromorphic vision sensor, adopts a new sampling paradigm, providing benefits including low power consumption, low latency, and high dynamic range~\cite{gallego2020event}.
As shown in Fig.~\ref{introduction}(i), unlike traditional RGB cameras that capture full video frames at fixed intervals, the event camera asynchronously records brightness changes at each pixel once the change exceeds a preset threshold.
This leads to higher temporal resolution and wider dynamic range, enabling accurate visual representation even in challenging scenarios.

However, due to its fundamentally different sampling mechanism, event data exhibits unique characteristics that differ significantly from conventional images.
As shown in Fig.~\ref{introduction}(ii), it is typically noisy and sparse, primarily containing motion-induced brightness variation signals, and thus overall represents limited and chaotic information with inherently deficient semantics, posing a formidable challenge for high-quality feature extraction.
Moreover, the scarcity of labeled event datasets has resulted in performance bottlenecks across various tasks.
Therefore, designing a self-supervised pre-training approach tailored to the physical properties of event data can better exploit their unique advantages and improve performance on downstream tasks without being constrained by the requirement for labeled data.

Currently, inspired by image-based self-supervised pre-training methods, some event-based approaches have been proposed and demonstrated competitive performance~\cite{yang2023event,yang2024event,klenk2024masked,ramesh2021zero,rolfe2016discrete,huang2024data}.
However, most existing event-based pre-training methods suffer from either noise sensitivity or inefficient utilization of event information, particularly in areas with sparser events.
Some methods~\cite{klenk2024masked, yang2024event} learn from noisy targets, while others~\cite{yang2023event, huang2024data} tackle sparsity and noise by neglecting or discarding sparser events.
For clearer analysis, as shown in Fig.~\ref{introduction}(iii), we divide event data into denser and sparser areas based on relative event density.
In denser regions, events are frequently triggered by significant light intensity changes, typically forming high-contrast edges, referred to as ``edge information''.
In sparser regions, events occur sparsely due to minimal intensity changes near the firing threshold, often reflecting low-contrast textures and being more susceptible to noise, referred to as ``texture cues''.
We term these two types of valuable information discoverable in event data as ``latent information''.
In our view, event data can be further refined to yield superior information with enhanced edges, while fine textures can still be revealed even in sparser and noisier areas.
Therefore, a key question arises: \emph{``How can we enhance event representation by fully revealing the latent information within the noisy and sparse event data?''}

In this paper, we propose an efficient event data pre-training framework that leverages the advantages of the image domain to fully reveal the latent information in events, thereby improving event representation.
As shown in Fig.~\ref{introduction}(a), images are clean and spatially dense, containing richer semantics that can compensate for the noisy and sparse nature of event data and enrich semantic content.
Although events and video frames differ in sampling paradigms, they can similarly represent intensity changes and fundamentally depict the same scene, enabling characteristic compensation for event data by exploring the intrinsic commonalities between event and image domains.
As illustrated in Fig.~\ref{introduction}(b), we first extract enhanced information of refined edges and fine textures, as presented in images, from the original event data through a difference-guided masked modeling stage that reconstructs temporal intensity difference maps built from images.
Then, since not all event information holds equal semantic value and its distribution is well-established in images, we improve semantic discrimination by employing focus-aimed contrastive learning that contrasts event and image features.
Additionally, to better preserve the abilities acquired from masked modeling and stabilize their effects on contrastive learning, we bridge the two stages using backbone-fixed feature transition method.
Our framework achieves superior performance across multiple downstream tasks, as shown in Fig.~\ref{introduction}(c).
In summary, our contributions are as follows:

1) We propose a novel physics-inspired self-supervised pre-training framework for noisy and sparse events that fully reveals the latent information within events by leveraging the commonalities between event and image domains.

2) We introduce temporal intensity difference maps to guide masked modeling for enhanced information extraction, and further incorporate feature transition and contrastive learning strategies to stabilize the learned representations and improve semantic discrimination.

3) Our method shows strong robustness to sparser and noisier events, consistently outperforming state-of-the-art approaches across multiple downstream tasks, including object recognition, semantic segmentation, and optical flow estimation.

\section{Related Works}
\label{sec:related_works}

\noindent\textbf{Image-based self-supervised learning.}
Self-supervised learning can be categorized into masked modeling, contrastive learning, and their combination.
Masked modeling predicts or reconstructs features or images by masking parts of the input image.
Early works predict patch features from dVAE~\cite{bao2021beit,zhou2021ibot}, while later works reconstruct features from momentum networks~\cite{baevski2022data2vec,tao2023siamese} or multi-modal pre-trained networks~\cite{wei2022mvp,hou2022milan}, and even directly reconstruct image pixels~\cite{he2022masked,gao2022convmae,xie2022simmim}.
Contrastive learning pulls together positive embeddings and pushes away negative ones, often using augmented image features~\cite{chen2020simple,he2020momentum} or multi-modal features~\cite{radford2021learning}, and can be extended to self-distillation~\cite{caron2020unsupervised,grill2020bootstrap}.
The combination of both methods, e.g., masking for contrastive learning~\cite{yi2022masked}, simultaneously reconstructing and contrasting features~\cite{huang2023contrastive,zhou2021ibot}, or contrastive learning after masked modeling~\cite{jiang2023layer}, has proven effective.
Unlike~\cite{jiang2023layer}, which employs layer grafting to resolve conflicts, we introduce a feature transition stage to bridge the two methods, aiming to stabilize the effect of masked modeling on contrastive learning.

\noindent\textbf{Event-based pre-training.}
Existing event-based pre-training methods all adopt self-supervised learning methods.
ECDP~\cite{yang2023event} masks sparser regions during pre-training, and contrasts the features of masked event images with those of their augmented counterparts and the corresponding RGB images.
It also proposes event feature regularization to avoid model collapse, and distribution alignment between event and image features to assist learning.
ECDDP~\cite{yang2024event} reconstructs features of masked event images using self-distillation, learning at three levels: patch-level, context-level, and image-level, and is particularly suitable for dense prediction tasks.
MEM~\cite{klenk2024masked} adapts BEIT~\cite{bao2021beit}'s approach to the event domain. It accumulates events into images and predicts the tokens of event image patches obtained in the dVAE~\cite{ramesh2021zero,rolfe2016discrete} stage.
Huang et al.~\cite{huang2024data} process denser events into voxel-level features and reconstruct them at both local and global levels, enabling the exploration of spatio-temporal cues and semantic information simultaneously.
As illustrated in Fig.~\ref{compare}, unlike ECDDP and MEM that learn from noisy targets, or ECDP and Huang et al. that ignore sparse regions, our method is robust to noise and can effectively extracts features from both dense and sparse areas, enhancing representation quality.

\noindent\textbf{Event-based downstream tasks.}
Event-based downstream tasks mainly include: global recognition (e.g., object recognition~\cite{gehrig2019end,kim2021n,messikommer2020event,bi2020graph,li2021graph,deng2022voxel,schaefer2022aegnn,xie2022vmv,deng2023dynamic,peng2023get}), dense prediction (e.g., semantic segmentation~\cite{alonso2019ev,sun2022ess}, optical flow estimation~\cite{gehrig2019end,wan2022learning}, depth estimation~\cite{hamaguchi2023hierarchical}, object detection~\cite{cannici2019asynchronous,messikommer2020event,li2021graph,schaefer2022aegnn,deng2023dynamic}), and sequence comprehension (e.g., action recognition~\cite{wu2021liaf,yao2021temporal,bi2020graph,deng2022voxel,xie2022vmv,deng2023dynamic,peng2023get}).
The focus of these three tasks varies, with different methods prioritizing specific tasks while remaining compatible with others.
For example, ECDP and MEM primarily target global recognition, ECDDP focuses on dense prediction, and Huang et al. emphasize sequence comprehension.
Our method is designed to extract spatial semantic features, making it well-suited for global recognition tasks that identify commonalities across objects.
It also enables precise inferences for dense prediction tasks by leveraging strong representations of edges, textures, and scene semantics.

\section{Method}
\label{sec:method}

\begin{figure*}[t]
\centering
\includegraphics[width=\textwidth]{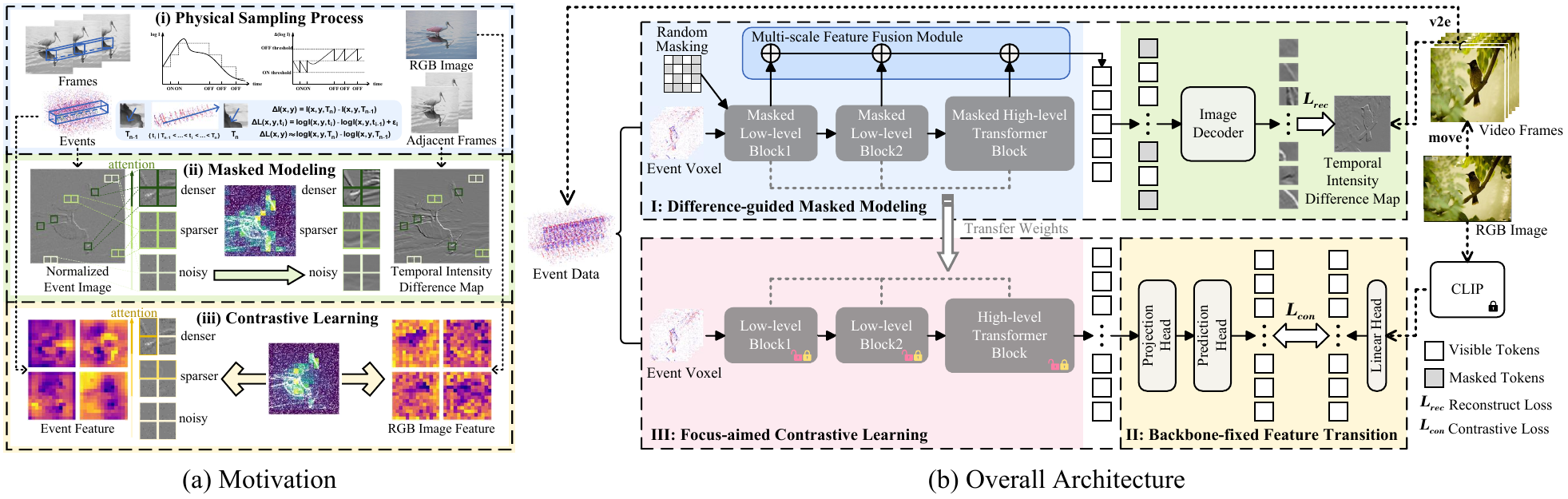}
\vspace{-8mm}
\caption{
The motivation and overall architecture of our method.
(a) Inspired by the physical sampling process of events, we use temporal intensity difference maps for masked modeling to extract enhanced information, and RGB image features for contrastive learning to improve semantic discrimination.
(b) Our method takes event voxels, temporal intensity difference maps, and RGB image features as input. The pre-training pipeline consists of three stages: Difference-guided Masked Modeling, Backbone-fixed Feature Transition, and Focus-aimed Contrastive Learning.
}
\vspace{-2mm}
\label{model}
\end{figure*}

\subsection{Motivation}
As previously discussed, our method aims to extract enhanced information and improve semantic discrimination from noisy and sparse events by leveraging the advantages of clean and dense images.
Accordingly, this section describes our motivation for determining learning strategies by considering the physical properties of events and uncovering their commonalities with images.

\noindent\textbf{Temporal intensity difference maps for masked modeling.}
To identify suitable learning targets in the image domain that correspond to and enhance the edge and texture information conveyed by events, we analyze the sampling processes of event data and video frames.
As shown in Fig.~\ref{model}(a)(i), video frames are typically sampled at fixed intervals, recording the light intensity $I$ at each pixel at specific timestamps.
Given two consecutive frames at $T_{n-1}$ and $T_n$, the intensity change at pixel $(x, y)$ over this interval can be expressed as:
\begin{equation}
\Delta I(x, y) = I(x, y, T_n) - I(x, y, T_{n-1}).
\label{eq_image}
\end{equation}
In contrast, between $T_{n-1}$ and $T_n$, the event camera continuously monitors the logarithmic intensity $L=log(I)$ at each pixel and triggers an event whenever the change exceeds a threshold $\theta$.
Multiple events may occur at timestamps $\{t_i \mid i=1,\dots,k, ; T_{n-1} < t_i < T_n\}$ if the threshold is crossed multiple times, formulated as:
\begin{equation}
\Delta L(x, y, t_i) = \log I(x, y, t_i) - \log I(x, y, t_{i-1}) + \epsilon_i = \theta,
\end{equation}
where $\epsilon_i$ represents the noise equivalent in intensity change.
While the frame-based intensity difference reflects the overall change in light intensity between $T_{n-1}$ and $T_n$, event cameras capture continuous intensity variations throughout the interval, accompanied by complex noise.
To approximate the total change, we accumulate the intensity variations induced by each event:
\begin{equation}
\Delta L(x, y) = \sum_{i=1}^k \Delta L(x, y, t_i) \approx \log I(x, y, T_n) - \log I(x, y, T_{n-1}),
\label{eq_event}
\end{equation}
Compared the Eq.\eqref{eq_image} and \eqref{eq_event}, we find that the intensity difference between two frames shares a similar physical significance with the accumulated inter-frame events.
This insight motivates us to use temporal intensity difference maps as ideal reconstruction targets for events, enabling information refinement via masked modeling.

As shown in Fig.~\ref{model}(a)(ii), the inter-frame events exhibit intensity patterns similar to their corresponding temporal intensity difference maps, which helps to comprehensively extract valid and enhanced information by strengthening prominent edges from denser events, detecting subtle textures from sparser events, and suppressing noise events.

\noindent\textbf{RGB image features for contrastive learning.}
To identify suitable learning targets with regular semantic distributions that effectively guide semantic learning, we observe that despite low-level differences arising from physical sampling mechanisms, events and images convey the same high-level semantics, with images offering richer details due to their higher spatial density.
Therefore, we leverage image features to guide semantic discrimination and use contrastive learning to further bridge the modality gap.

As shown in Fig.~\ref{model}(a)(iii), to better align with paired image features, the model can effectively identify and focus on regions with higher semantic value, such as the flamingo’s head, while capturing fine-grained textures like its wings and avoiding irrelevant edges and noisy background areas.
 
\subsection{Overall architecture}
As shown in Fig.~\ref{model}(b), our event-based self-supervised pre-training framework takes three types of paired inputs: event voxels, temporal intensity difference maps, and RGB image features.
To fully exploit the latent information in noisy and sparse events, the framework comprises three stages:
\textbf{i) Difference-guided masked modeling.}
Reconstruct masked patches in temporal difference maps to learn fine-grained edge and texture information from events.
\textbf{ii) Backbone-fixed feature transition.}
Contrast event and image features while only updating the contrastive heads, serving as a bridge between stages.
\textbf{iii) Focus-aimed contrastive learning.}
Train the entire network with the same contrastive strategy as before to learn semantics and better distinguish valuable regions.

\subsection{Pre-training data preparation}
\label{data_preparation}
Since strict spatio-temporal alignment between events and reconstructed targets is crucial for masked modeling, we simulate synchronous videos and event sequences that naturally provide such alignment.
This approach addresses the current lack of event pre-training datasets meeting these requirements.
Compared to recording images displayed on a monitor with a moving event camera, it is more cost-effective, supports a wider range of motion patterns, and avoids interference from external noise.
Specifically, we generate paired data from a single RGB image as follows:
\textbf{i) Event voxels.}
We apply random combinations of motion modes to an RGB image $I$ to generate simulated video frames $F_{1,\dots,n}^{rgb}$.
The corresponding event sequence $E$ with realistic noise is generated using v2e~\cite{hu2021v2e}, and segmented into $\varepsilon_{1,\dots,n-1}$ aligned with frame intervals.
Each segment is converted into an event voxel $V_{1,\dots,n-1}^\varepsilon \in \mathbb{R}^{H \times W \times b}$, where $H \times W$ is the spatial resolution and $b$ is the number of time bins, serving as input that preserves motion and temporal cues.
\textbf{ii) Temporal intensity difference maps.}
The simulated RGB frames are converted into grayscale images $F_{1,\dots,n}^{gray}$.
Temporal intensity difference maps $F_{1,\dots,n-1}^d$ are computed by subtracting adjacent frames:
$F_i^d = F_{i+1}^{gray} - F_i^{gray}, \ i = 1,\dots,n-1$,
ensuring alignment with the corresponding event voxel $V_i^\varepsilon$.
These maps serve as reconstruction targets during masked modeling.
\textbf{iii) RGB image features.}
RGB image features $f^{rgb}$ are extracted by feeding the RGB image into a pre-trained CLIP~\cite{radford2021learning} network:
$f^{rgb} = {\rm CLIP}(I)$.
These features serve as contrastive targets during contrastive learning.

\subsection{Difference-guided masked modeling}
The first stage establishes the foundation for the entire event representation learning process.

As shown in Fig.~\ref{model}(b)(I), we use a Transformer-based backbone, including ViT~\cite{dosovitskiy2020image}, ConvViT~\cite{gao2022convmae}, and MAE-compatible Swin-Transformer~\cite{huang2022green}, with an encoder-decoder structure.
The event voxel $V^\varepsilon$ is divided into patches and randomly masked to balance event density between visible and masked patches, preserving both dense and sparse regions.
A multi-scale feature fusion module combines low-level and high-level features to aid texture detail reconstruction while maintaining high-level semantics from deeper layers.
For the ViT, features from the 2$^{nd}$, 4$^{th}$, and top layers are fused.
For ConvViT and Swin-Transformer, features from the first two layers, processed with stride convolutions as in~\cite{gao2022convmae}, are fused with those from the top Transformer block.
The feature tokens extracted by the encoder from visible patches are represented as:
\begin{equation}
T_v^\varepsilon=E_l\left(\mathcal{P}_v^\varepsilon\right)+E_h\left(\mathcal{P}_v^\varepsilon\right),
\end{equation}
where $\mathcal{P}_v^\varepsilon$ denotes the visible patches of $V^\varepsilon$, and $E_l$, $E_h$ are the encoder’s low-level and high-level blocks, respectively.
Masked tokens $T_m^\varepsilon$ are then added and, together with visible tokens $T_v^\varepsilon$, fed into the Transformer-based decoder $D$ to reconstruct the invisible patches of temporal intensity difference maps:
$F^{rec} = D\left(T^\varepsilon\right)$, where $T^\varepsilon=\{T^\varepsilon_v, T^\varepsilon_m\}$.

We perform masked modeling to reconstruct masked patches of the temporal intensity difference map from visible event data and compute the MSE loss against the standardized target patches as follows:
\begin{equation}\label{L_rec}
\mathcal{L}_{rec}=\frac{1}{n}\sum_{i=1}^n\left(F^{rec}_{m,i}-F^{ds}_{m,i}\right)^2,
\end{equation}
where $n$ is the number of pixels in masked patches,
$\{F^{rec}_{m,i}|i=1,\dots,n\}$ are the reconstructed pixel values of masked patches,
and $\{F^{ds}_{m,i}|i=1,\dots,n\}$ are the standardized pixels of target masked temporal intensity difference map pixels $\{F^d_{m,i}|i=1,\dots,n\}$, calculated as:
$F_{m,i}^{ds}=\frac{F^d_{m,i}-\mu}{\sigma}$,
where $\mu=\frac{1}{n}\sum_{i=1}^nF^d_{m,i}$ and $\sigma=\sqrt{\frac{1}{n}\sum_{i=1}^n\left(F^d_{m,i}-\mu \right)^2}$.

\begin{figure}[t]
\centering
\includegraphics[width=\columnwidth]{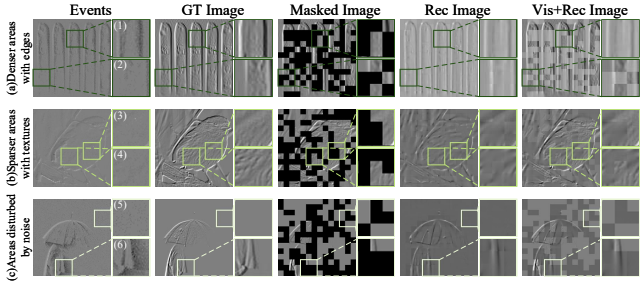}
\vspace{-8mm}
\caption{Masked modeling result.
(a) Denser areas with edges: (1) reconstruct clear edges, and (2) fill in missing edges.
(b) Sparser areas with textures: (2) and (3) infer subtle textures from minimal intensity difference.
(c) Areas disturbed by noise: (4) remove noise caused by motion blur, and (5) the inherent noise of camera.
Color differences in reconstructed patches result from the normalization of temporal intensity difference maps in the masked modeling loss (Eq.~\eqref{L_rec}).
}
\vspace{-4mm}
\label{pr_rec}
\end{figure}

\subsection{Backbone-fixed feature transition}
The second stage bridges the previous and next stages, preserving the abilities learned from masked modeling and stabilizing their impact on contrastive learning.

As shown in Fig.~\ref{model}(b)(II), in this stage, we remove the multi-scale feature fusion module and transfer the backbone weights from the masked modeling stage.
By focusing solely on high-level features, it also reduces conflicts between low-level reconstruction and high-level feature comparison.
Contrastive heads are added and updated during training, while the backbone network weights remain frozen.
The fixed backbone embeds the event voxel $V^\varepsilon$ into $f_x^\varepsilon$, which is then projected to $f_{xp}^\varepsilon$ through projection and prediction heads.
Meanwhile, the RGB image feature $f^{rgb}$ is projected to $f_p^{rgb}$ via a linear head.

We perform contrastive learning between event and image features and compute the loss $\mathcal{L}_{con}\left(f_{xp}^\varepsilon,\{f_{p}^{rgb}\}\right)$ using infoNCE formulation~\cite{oord2018representation} $\mathcal{L}_{con}\left(q,\{k\}\right)$: 
\begin{equation}
\mathcal{L}_{con}=-\log\frac{\exp\left(q \cdot {k_+}/{\tau}\right)}{\exp\left(q \cdot {k_+}/{\tau}\right)+\sum\limits_{k_-}\exp\left(q \cdot {k_-}/{\tau}\right)},
\end{equation}
where $q=f_{xp}^\varepsilon$ is the projected event feature, $k_+=f_{p}^{rgb}$ is the paired projected RGB image feature, $k_-$ are the projected unpaired RGB features stored in queue, and $\tau$ is the temperature hyperparameter.

\subsection{Focus-aimed contrastive learning}\label{sec_cl}

Equipped with the backbone network’s information extraction ability from the masked modeling stage and the preliminary feature projection capability of the contrastive head from the feature transition stage, the last stage ultimately achieves strong event representational ability through efficient training.

As shown in Fig.~\ref{model}(b)(III), we adopt the same contrastive learning method as in the feature transition stage but update the entire network, including the backbone.
During training, we compute the contrastive loss $\mathcal{L}_{con}\left(f_p^\varepsilon,\{f_{p}^{rgb}\}\right)$, where $f_p^\varepsilon$ denotes the projected event features from the unfrozen backbone network.


\section{Experiments}
\label{sec:experiments}

\begin{figure}[t]
\centering
\includegraphics[width=\columnwidth]{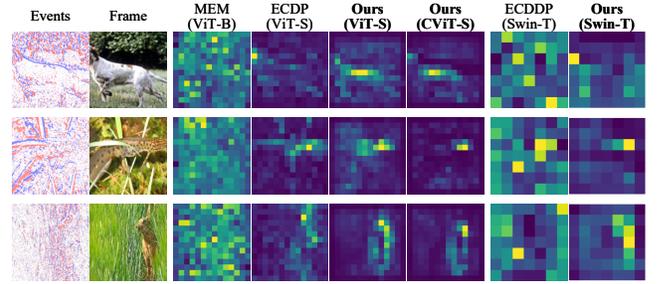}
\vspace{-8mm}
\caption{
Attention map comparison of different methods.
Our method is able to accurately focus on entire semantic foregrounds while mitigating the impact of cluttered backgrounds.
}
\vspace{-4mm}
\label{pr_attn}
\end{figure}

\begin{table*}[t]
\scriptsize
\caption{Comparison of the performance on object recognition tasks using multiple small-scale datasets.}
\vspace{-3mm}
\label{obj_rec_s}
\setlength\tabcolsep{0.1cm}
\belowrulesep=0pt
\aboverulesep=0pt
\renewcommand{\arraystretch}{0.7}
\begin{threeparttable}
{
\begin{tabular*}{\linewidth}{>
{\centering\arraybackslash}p{2.935cm}|>
{\centering\arraybackslash}p{1.8cm}>
{\centering\arraybackslash}p{1.0cm}|>
{\centering\arraybackslash}p{1.8cm}>
{\centering\arraybackslash}p{1.0cm}>
{\centering\arraybackslash}p{1.4cm}|>
{\centering\arraybackslash}p{2.0cm}>
{\centering\arraybackslash}p{2.0cm}>
{\centering\arraybackslash}p{2.0cm}}
\toprule
\multirow{2}{*}{Method} & \multirow{2}{*}{Backbone} & \multirow{2}{*}{Params} & \multicolumn{3}{c|}{Pre-training} & \multicolumn{3}{c}{Top-1 Acc(\%)} \\
\cline{4-9}
 &  &  & Dataset & Epochs & Training time(h) & N-Caltech101 & CIFAR10-DVS & N-Cars \\
\midrule
\multicolumn{9}{l}{Task-specific event-based methods.} \\
\midrule
N-ImageNet~\cite{kim2021n} & - & - & - & - & - & \textbf{86.81} & 73.72 & 94.73 \\ 
AsyNet~\cite{messikommer2020event} & - & - & - & - & - & 74.5 & 66.3 & 94.4 \\
RG-CNNs~\cite{bi2020graph} & - & - & - & - & - & 65.7 & 54.0 & 91.4 \\
EvS-S~\cite{li2021graph} & - & - & - & - & - & 76.1 & 68.0 & 93.1 \\
EV-VGCNN~\cite{deng2022voxel} & - & - & - & - & - & 74.8 & 67.0 & 95.3 \\
AEGNN~\cite{schaefer2022aegnn} & - & - & - & - & - & 66.8 & - & 94.5 \\
VMV-GCN~\cite{xie2022vmv} & - & - & - & - & - & 77.8 & 69.0 & 93.2 \\
EDGCN~\cite{deng2023dynamic} & - & - & - & - & - & 80.1 & 71.6 & 95.8 \\
GET~\cite{peng2023get} & - & - & - & - & - & - & \textbf{78.1} & \textbf{96.7} \\
\midrule
\multicolumn{9}{l}{Backbone training from scratch.} \\
\midrule
Huang et al.~\cite{huang2024data} & light ViT-S & 2.2M & - & - & - & 81.4 & 71.0 & \textbf{96.4} \\
Huang et al.~\cite{huang2024data} & light ViT-B & 13.5M & - & - & - & \textbf{83.4} & 71.9 & 95.9 \\
MEM~\cite{klenk2024masked} & ViT-B/16 & 85.8M & - & - & - & 66.94 & - & 92.71 \\
ECDP~\cite{yang2023event} & ResNet50 & 23.0M & - & - & - & 62.69 & 56.65 & 91.20 \\
ECDP~\cite{yang2023event} & ViT-S/16 & 21.6M & - & - & - & 55.63 & 52.45 & 89.14 \\
Ours & ViT-S/16 & 21.8M & - & - & - & 67.77 & 67.99 & 94.12 \\
Ours & ConvViT-S/16 & 21.9M & - & - & - & 70.60 & \textbf{72.73} & 95.92 \\
Ours & Swin-T/7 & 27.5M & - & - & - & 68.40 & 68.40 & 95.04 \\
\midrule
\multicolumn{9}{l}{Event-based self-supervised pre-training methods.} \\
\midrule
Huang et al.~\cite{huang2024data} & light ViT-S & 2.2M & N-Caltech101 & 700 & - & 86.0$^*$ & 75.9 & 97.1 \\
Huang et al.~\cite{huang2024data} & light ViT-B & 13.5M & N-Caltech101 & 700 & - & 88.0$^*$ & 78.6 & 97.1 \\
MEM~\cite{klenk2024masked} & ViT-B/16 & 85.8M & N-Cars & 1000 & 12 & - & - & \textbf{98.55}$^*$ \\
MEM~\cite{klenk2024masked} & ViT-B/16 & 85.8M & N-Caltech101 & 3000 & 58 & 85.60$^*$ & - & - \\
MEM~\cite{klenk2024masked} & ViT-B/16 & 85.8M & N-ImageNet & 75 & 275 & 90.10 & - & 93.27 \\
ECDP~\cite{yang2023event} & ResNet50 & 23.0M & N-ImageNet$^+$ & 300 & 267 & 87.08 & 74.75 & 98.01 \\
ECDP~\cite{yang2023event} & ViT-S/16 & 21.6M & N-Caltech101 & 3000 & 14 & 85.4$^*$ & 76.9 & 95.0 \\
ECDP~\cite{yang2023event} & ViT-S/16 & 21.6M & N-ImageNet$^+$ & 300 & 267 & 87.66 & 78.00 & 97.93 \\
ECDDP~\cite{yang2024event} & Swin-T/7 & 27.5M & E-TartanAir & 300 & 152 & 82.78 & 82.39 & 97.15 \\
\rowcolor{gray!20}Ours & ViT-S/16 & 21.8M & EF-ImageNet$^+$ & 710 & 80 & 90.45 & \textbf{84.93} & 97.74 \\
\rowcolor{gray!20}Ours & ConvViT-S/16 & 21.9M & EF-ImageNet$^+$ & 710 & 88 & \textbf{92.36} & \underline{84.53} & \underline{98.29} \\
\rowcolor{gray!20}Ours & Swin-T/7 & 27.5M & EF-ImageNet$^+$ & 710 & 91 & \underline{91.03} & 84.17 & 98.19 \\
\bottomrule
\end{tabular*}}
\begin{tablenotes}[para,flushleft]
`+': method using paired RGB images with events.
`*': method fine-tuning on the same dataset as pre-training, resulting in higher performance on this dataset.
Data marked with \textbf{bold} and \underline{underline}: the best and the second-best performance in the group.
\end{tablenotes}
\end{threeparttable}
\vspace{-4mm}
\end{table*}

\begin{table}
\scriptsize
\caption{Comparison of the performance on object recognition tasks using large-scale datasets.}
\vspace{-3mm}
\label{obj_rec_img}
\setlength\tabcolsep{0.14cm}
\belowrulesep=0pt
\aboverulesep=0pt
\renewcommand{\arraystretch}{0.7}
\begin{threeparttable}
{
\begin{tabular}{>
{\centering\arraybackslash}p{1.43cm}|>
{\centering\arraybackslash}p{0.8cm}|>
{\centering\arraybackslash}p{0.77cm}>
{\centering\arraybackslash}p{0.45cm}|>
{\centering\arraybackslash}p{0.63cm}>
{\centering\arraybackslash}p{0.63cm}|>
{\centering\arraybackslash}p{0.63cm}|>
{\centering\arraybackslash}p{0.63cm}}
\toprule
\multirow{3}{*}{Method} & \multirow{3}{*}{Backbone} & \multicolumn{2}{c|}{Pre-training} & \multicolumn{4}{c}{Top-1 Acc(\%)} \\
\cline{3-8}
 &  & \multirow{2}{*}{Dataset} & \multirow{2}{*}{Time} & \multicolumn{3}{c|}{Mini-N} & \multirow{2}{*}{Mini-ES} \\
 \cline{5-7}
 &  &  &  & Org & Var & Mean & \\
\midrule
\multicolumn{6}{l}{Backbone training from scratch.} \\
\midrule
Ours & ViT-S & - & - & 33.95 & 29.36 & 29.82 & 32.50 \\
Ours & CViT-S & - & - & 45.53 & 36.01 & 36.97 & \textbf{51.81} \\
Ours & Swin-T & - & - & \textbf{47.64} & \textbf{40.45} & \textbf{41.17} & 38.72 \\
\midrule
\multicolumn{6}{l}{Event-based self-supervised pre-training methods.} \\
\midrule
MEM~\cite{klenk2024masked} & ViT-B & N-Img$^+$ & 275 & 65.19$^\checkmark$ & 54.36$^\circ$ & 55.44$^*$ & 54.75 \\
ECDP~\cite{yang2023event} & ViT-S & N-Img$^+$ & 267 & \textbf{72.35}$^\checkmark$ & \textbf{62.23}$^\circ$ & \textbf{63.25}$^*$ & \textbf{66.02} \\
\cline{1-7}
ECDP\&~\cite{yang2023event} & ViT-S & EF-Img$^+$ & 63 & 41.81$^\times$ & 40.09$^\times$ & 40.26 & 49.96 \\
\rowcolor{gray!20}Ours & ViT-S & EF-Img$^+$ & 80 & \textbf{46.67}$^\times$ & \textbf{47.95}$^\times$ & \textbf{47.82} & 60.17 \\
\cline{1-7}
\rowcolor{gray!20}Ours & ViT-S & EF+N$^+$ & 97 & 50.26$^{\times\checkmark}$ & 52.95$^{\times\circ}$ & 52.68$^{(*)}$ & \underline{63.96} \\
\midrule
ECDDP~\cite{yang2024event} & Swin-T & E-Tar & 152 & 46.50$^\times$ & 44.09$^\times$ & 44.33 & 62.77 \\
\rowcolor{gray!20}Ours & CViT-S & EF-Img$^+$ & 88 & \textbf{58.07}$^\times$ & \textbf{55.55}$^\times$ & \textbf{55.80} & \underline{66.86} \\
\rowcolor{gray!20}Ours & Swin-T & EF-Img$^+$ & 91 & \underline{51.06}$^\times$ 
 & \underline{50.71}$^\times$ & \underline{50.75} & 64.10 \\
\cline{1-7}
\rowcolor{gray!20}Ours & CViT-S & EF+N$^+$ & 109 & 60.08$^{\times\checkmark}$ & 61.53$^{\times\circ}$ & 61.38$^{(*)}$ & \textbf{71.16} \\
\bottomrule
\end{tabular}}
\begin{tablenotes}[para,flushleft]
The indications of `+', `*', \textbf{bold} and \underline{underline} are the same as those in Table~\ref{obj_rec_s}.
`(*)': method using partly overlapping datasets for pre-training and fine-tuning.
`\&': method retrained with the same pre-training datasets as ours.
`Org', `Var' and `Mean': performance on the original test set, average performance over 9 variant test sets (including 5 motion trajectories and 4 brightness conditions), and overall average across all 10 N-ImageNet test sets, respectively.
`$\checkmark$', `$\times$' and `$\circ$' in the `Org' and `Var' columns further indicate whether the test datasets are the same as, different from, or partially overlapping with the pre-training datasets, respectively.
\end{tablenotes}
\end{threeparttable}
\vspace{-4mm}
\end{table}

\subsection{Experimental setup}
\noindent\textbf{Implementation.}
We experiment with our method using three Transformer-based backbone networks: ViT~\cite{dosovitskiy2020image}, ConvViT~\cite{gao2022convmae} and MAE-compatible Swin-Transformer~\cite{huang2022green}.
For masked modeling, we set the masking ratio to 50\% and use an 8-layer and 512-width Transformer as the image decoder.
For contrastive learning, we use 3-layer and 2-layer MLP heads as projection and prediction heads respectively, followed by MoCo v3~\cite{chen2021empirical}, a pre-trained ViT-B/16 CLIP~\cite{radford2021learning} network for RGB image feature extraction, and a queue of length 1024 for storing negative samples.
We append UperNet~\cite{xiao2018unified} and FCN~\cite{long2015fully} as decode head and auxiliary head, respectively, for dense prediction tasks.

\noindent\textbf{Pre-training dataset.}
As stated in Sec.~\ref{data_preparation}, to meet the data demands of our pre-training method, we simulate an event-frame alignment pre-training dataset (EF-ImageNet) using the top 10\% of each class in ImageNet-1K~\cite{deng2009imagenet}.
More details are in the Supp.

\noindent\textbf{Baseline.}
i) Task-specific event-based methods: conduct supervised training using their own network structures.
ii) Backbone training from scratch: conduct supervised training with randomly initialized weights using different backbones.
Since training from scratch performs much worse than pre-training and is more sensitive to factors like initialization and training configurations, it is only used for comparison.
iii) Event-based self-supervised pre-training methods: conduct self-supervised pre-training on event datasets and then fine-tune.
For dense prediction tasks, due to the clear performance gap between different backbones (observable in scratch training), we group them into two categories: original ViT and multi-scale ViT (including ConvViT and Swin-Transformer).
Additionally, since fine-tuning strategies vary across methods (e.g., ECDP~\cite{yang2023event} uses the architecture from~\cite{yue2021vision}; ECDDP~\cite{yang2024event} applies TTA, TMA~\cite{liu2023tma}, and intermediate fine-tuning), we ensure a fair comparison by fine-tuning all methods with available pretrained weights, using the same base architectures and training configurations.

\noindent\textbf{Pre-training time.}
We evaluate the training time of all pre-training methods on a single NVIDIA GeForce RTX 3090 GPU with a batch size of 64 to assess training efficiency.
Additional training cost evaluations are in the Supp.

\subsection{Pre-training model analysis}\label{pr_analysis}

\noindent\textbf{Masked modeling results.}
Fig.~\ref{pr_rec}(a) shows that our model reconstructs clear, continuous edges using intensity and orientation information from dense events, achieving quality comparable to temporal intensity difference maps.
Fig.~\ref{pr_rec}(b) illustrates that the model infers subtle textures in sparse regions by leveraging surrounding textures and semantic cues.
Fig.~\ref{pr_rec}(c) demonstrates the model's ability to distinguish valid events from noise, eliminating motion blur and inherent noise.
These results highlight the effectiveness of our masked modeling method in revealing edge and texture information from raw events, laying a solid foundation for better representation learning.

\noindent\textbf{Attention maps.}
Fig.~\ref{pr_attn} shows a comparison of attention maps from different methods.
The attention maps of MEM~\cite{klenk2024masked} and ECDDP lack clear patterns.
ECDP~\cite{yang2023event} generally focuses on semantic areas but tends to prioritize denser events, which may limit its ability to capture complete object details and make it more susceptible to irrelevant events and noise.
In contrast, our method accurately focuses on the entire semantic foreground, such as the dog’s head and body, while mitigating the impact of cluttered backgrounds like grass, leading to improved event representation quality.

\begin{figure*}[t]
\centering
\includegraphics[width=\textwidth]{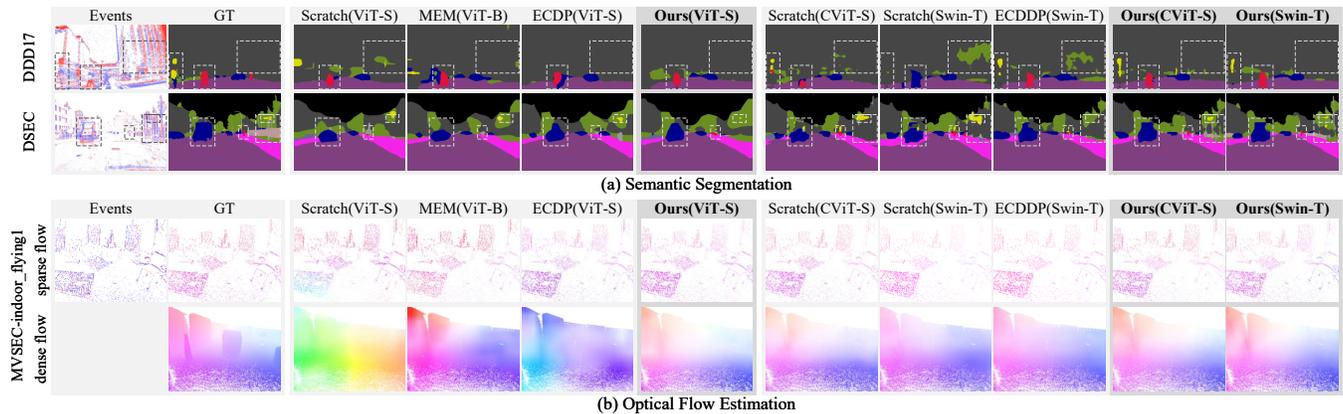}
\vspace{-8mm}
\caption{
Qualitative comparison on semantic segmentation and optical flow estimation.
(a) Our method helps segment indistinct objects, provide more accurate edges, and can avoid incorrect segmentation.
(b) Our method can predict the magnitude and direction of the optical flow vector more precisely.
}
\label{ft_result}
\vspace{-3mm}
\end{figure*}

\subsection{Performance on downstream tasks}\label{down_exp}
\noindent\textbf{Object recognition on multiple small-scale datasets.}
We evaluate our method on three small-scale datasets: N-Caltech101~\cite{orchard2015converting}, CIFAR10-DVS~\cite{li2017cifar10}, and N-Cars~\cite{sironi2018hats}.
As shown in Table~\ref{obj_rec_s}, our method outperforms all event-based and training-from-scratch approaches, and surpasses most event-based self-supervised pre-training methods with relatively short training time.
Notably, compared to ECDP, which also uses a ViT-S/16 backbone and paired RGB data, our ViT-based method achieves 2.79\% and 6.93\% higher accuracy on N-Caltech101 and CIFAR10-DVS, respectively, with less pre-training time.
Using both ConvViT and Swin-Transformer backbones, our method achieves similarly strong or superior performance.
Although the N-Cars dataset poses challenges due to its real-world data gap, our ConvViT-based method still achieves the second-highest performance, only behind MEM (which is pre-trained and fine-tuned on the same dataset), while our ViT-based method matches ECDP.
We also evaluate the performance of ECDDP and observe suboptimal and unstable results, while our Swin-based method demonstrates better performance.

\begin{table}[t]
\scriptsize
\centering
\caption{Comparison of the performance on semantic segmentation tasks.}
\vspace{-3mm}
\label{sem_seg}
\setlength\tabcolsep{0.1cm}
\belowrulesep=0pt
\aboverulesep=0pt
\begin{threeparttable}
\renewcommand{\arraystretch}{0.7}
\begin{tabularx}{\linewidth}{>
{\centering\arraybackslash}p{1.5cm}|>
{\centering\arraybackslash}p{1.0cm}|>
{\centering\arraybackslash}p{1.0cm}>
{\centering\arraybackslash}p{0.7cm}>
{\centering\arraybackslash}p{0.45cm}|>
{\centering\arraybackslash}p{1.18cm}>
{\centering\arraybackslash}p{1.19cm}}
\toprule
\multirow{2}{*}{Method} & \multirow{2}{*}{Backbone} & \multicolumn{3}{c|}{Pre-training} & \multicolumn{2}{c}{mIOU(\%)} \\
\cline{3-7}
 & & Dataset & Epochs & Time & DDD17 & DSEC \\
\midrule
\multicolumn{7}{l}{Task-specific event-based methods.} \\
\midrule
EV-SegNet~\cite{alonso2019ev} & - & - & - & - & 54.81 & 51.76 \\
ESS~\cite{sun2022ess} & - & - & - & - & \textbf{61.37}$^+$ & \textbf{53.29}$^+$ \\
\midrule
\multicolumn{7}{l}{Backbone training from scratch.} \\
\midrule
MEM~\cite{klenk2024masked} & ViT-B & - & - & - & - & 42.79 \\
ECDP~\cite{yang2023event} & ViT-S & - & - & - & 43.89 & 38.24 \\
Ours & ViT-S & - & - & - & 49.95 & 39.78 \\
Ours & CViT-S & - & - & - & \textbf{52.01} & \textbf{45.84} \\
Ours & Swin-T & - & - & - & 51.32 & 45.49 \\
\midrule
\multicolumn{7}{l}{Event-based self-supervised pre-training methods.} \\
\midrule
MEM~\cite{klenk2024masked} & ViT-B & DSEC & 6000 & 70 & - & 44.62$^*$ \\
ECDP~\cite{yang2023event} & ViT-S & N-Img$^+$ & 300 & 267 & 54.66 & 47.91 \\
\midrule
MEM\#~\cite{klenk2024masked} & ViT-B & N-Img & 75 & 275 & 53.65 & 46.83 \\
ECDP\#~\cite{yang2023event} & ViT-S & N-Img$^+$ & 300 & 267 & \underline{54.49} & \textbf{47.56} \\
\rowcolor{gray!20}Ours & ViT-S & EF-Img$^+$ & 710 & 80 & \textbf{56.06} & \underline{47.24} \\
\midrule
ECDDP\#~\cite{yang2024event} & Swin-T & E-Tar & 300 & 152 & 59.43 & 50.43 \\
\rowcolor{gray!20}Ours & CViT-S & EF-Img$^+$ & 710 & 88 & \underline{60.53} & \textbf{53.22} \\
\rowcolor{gray!20}Ours & Swin-T & EF-Img$^+$ & 710 & 91 & \textbf{61.45} & \underline{53.16} \\
\bottomrule
\end{tabularx}
\begin{tablenotes}[para,flushleft]
The indications of `+', `*', \textbf{bold} and \underline{underline} are the same as those in Table~\ref{obj_rec_s}.
`\#': method fine-tuning using consistent base architectures and training method as ours.
\end{tablenotes}
\end{threeparttable}
\vspace{-4mm}
\end{table}

\noindent\textbf{Object recognition on large-scale datasets.}
We evaluate two large-scale real-world event datasets, N-ImageNet~\cite{kim2021n} and ES-ImageNet~\cite{lin2021imagenet}, using their mini versions consisting of the first 100 classes.
As shown in Table~\ref{obj_rec_img}, on N-ImageNet, ECDP and MEM—both pre-trained and fine-tuned on the same datasets—exhibit superior performance on the original test set but suffer a notable drop on the variants.
This suggests using the same datasets narrows the gap between pre-training and fine-tuning by reducing differences in training scenes and event distributions from sampling configurations, but may confound evaluation of pre-training effectiveness.
To ensure a fair comparison, we retrain ECDP using the same pre-training datasets as ours.
Our method outperforms it across the `Org', `Var' and `Mean' test sets.
Through contrastive learning on N-ImageNet, our method achieves additional performance improvements.
On ES-ImageNet, our method also delivers most of the highest performance among all methods.
Our ViT-based method ranks surpass the retrained ECDP under the same pre-training conditions.
These results demonstrate the strong object representation ability of our method.

\begin{table}[t]
\scriptsize
\centering
\caption{Comparison of the performance on optical flow estimation tasks.}
\vspace{-3mm}
\label{opt_flow_est}
\setlength\tabcolsep{0.1cm}
\belowrulesep=0pt
\aboverulesep=0pt
\begin{threeparttable}
\renewcommand{\arraystretch}{0.7}
\begin{tabularx}{\columnwidth}{>
{\centering\arraybackslash}p{1.5cm}|>
{\centering\arraybackslash}p{1.0cm}|>
{\centering\arraybackslash}p{1.0cm}>
{\centering\arraybackslash}p{0.45cm}|>
{\centering\arraybackslash}p{1.02cm}>
{\centering\arraybackslash}p{1.02cm}>
{\centering\arraybackslash}p{1.03cm}}
\toprule
\multirow{2}{*}{Method} & \multirow{2}{*}{Backbone} & \multicolumn{2}{c|}{Pre-training} & \multicolumn{3}{c}{AEE} \\
\cline{3-7}
 &  & Dataset & Time & indoor1 & indoor2 & indoor3 \\
\midrule
\multicolumn{7}{l}{Task-specific event-based methods.} \\
\midrule
EST~\cite{gehrig2019end} & - & - & - & 1.24 & 2.05 & 1.71 \\
DCEIFlow~\cite{wan2022learning} & - & - & - & \textbf{0.748} & \textbf{1.388} & \textbf{1.132} \\
\midrule
\multicolumn{7}{l}{Backbone training from scratch.} \\
\midrule
ECDP~\cite{yang2023event} & ViT-S & - & - & 0.64 & 1.36 & 1.05 \\
Ours & ViT-S & - & - & 0.734 & 1.630 & 1.292 \\
Ours & CViT-S & - & - & 0.608 & 1.384 & 0.996 \\
Ours & Swin-T & - & - & \textbf{0.491} & \textbf{1.110} & \textbf{0.804} \\
\midrule
\multicolumn{7}{l}{Event-based self-supervised pre-training methods.} \\
\midrule
ECDP~\cite{yang2023event} & ViT-S & N-Img$^+$ & 267 & 0.614 & 1.261 & 1.001 \\
\midrule
MEM\#~\cite{klenk2024masked} & ViT-B & N-Img & 275 & 0.605 & 1.577 & 1.122 \\
ECDP\#~\cite{yang2023event} & ViT-S & N-Img$^+$ & 267 & \underline{0.561} & \underline{1.496} & \underline{0.985} \\
\rowcolor{gray!20}Ours & ViT-S & EF-Img$^+$ & 80 & \textbf{0.457} & \textbf{1.076} & \textbf{0.768} \\
\midrule
ECDDP\#~\cite{yang2024event} & Swin-T & E-Tar & 152 & \underline{0.432} & \textbf{0.904} & \textbf{0.638} \\
\rowcolor{gray!20}Ours & CViT-S & EF-Img$^+$ & 88 & 0.434 & 1.035 & 0.778 \\
\rowcolor{gray!20}Ours & Swin-T & EF-Img$^+$ & 91 & \textbf{0.416} & \underline{1.002} & \underline{0.742} \\
\bottomrule
\end{tabularx}
\begin{tablenotes}[para,flushleft]
The indications of `+', `*', `()', \textbf{bold} and \underline{underline} are the same as those in Table~\ref{obj_rec_s}.
The indications of `\#' is the same as those in Table~\ref{sem_seg}.
\end{tablenotes}
\end{threeparttable}
\vspace{-4mm}
\end{table}

\noindent\textbf{Semantic segmentation.}
We implement semantic segmentation on the DSEC~\cite{gehrig2021dsec} and DDD17~\cite{binas2017ddd17} datasets as shown in Table~\ref{sem_seg}.
Our method achieves most of the highest performance among event-based pre-training methods in both original ViT and multi-scale ViT groups.
Fig.~\ref{ft_result}(a) shows our method effectively segments indistinct objects like people and poles, enhances edge accuracy for cars, and avoids missegmenting trees.
This demonstrates its ability to identify semantic regions and accurately locate clear edges.

\noindent\textbf{Optical flow estimation.}
We show our optical flow estimation results on the MVSEC dataset in Table~\ref{opt_flow_est}.
Our method also achieves most of the highest performance among event-based pre-training methods in both original ViT and multi-scale ViT groups.
The Swin-based method is slightly worse than ECDDP on indoor\_flying2 and indoor\_flying3, but ECDDP focuses on dense prediction and performs worse on global recognition compared to ours.
Fig.~\ref{ft_result}(b) shows our method predicts optical flow magnitude and direction more accurately, benefiting from efficient latent information extraction from events across all regions.

\subsection{Robustness study}
We evaluate the robustness of our method against ECDP, MEM, and ECDDP on N-Caltech101, ES-ImageNet and N-Imagenet.
For N-Caltech101 and ES-ImageNet, we manually construct three test variants: i) Sparse: fewer events, ii) Noise: added or removed events, and iii) Sparse and Noise.
For N-ImageNet, we use its provided variants covering five motion types and four brightness levels.
We compute the relative performance gap between the original and variant test sets.
As shown in Table~\ref{robust}, our methods achieve the smallest performance difference in most cases, demonstrating robustness in handling sparsity and noise issues in event data, as well as variations in edge and texture information arising from different motion patterns and brightness conditions.

\begin{table}[t]
\scriptsize
\caption{Robustness study on object recognition tasks.}
\vspace{-3mm}
\label{robust}
\setlength\tabcolsep{0.1cm}
\belowrulesep=0pt
\aboverulesep=0pt
\begin{threeparttable}
\renewcommand{\arraystretch}{0.7}
\begin{tabularx}{\linewidth}{>
{\centering\arraybackslash}p{0.8cm}|>
{\centering\arraybackslash}p{0.8cm}|>
{\centering\arraybackslash}p{1.1cm}|>
{\centering\arraybackslash}p{1.5cm}>
{\centering\arraybackslash}p{1.5cm}>
{\centering\arraybackslash}p{1.52cm}}
\toprule
\multirow{2}{*}{\makecell[c]{Method}} & \multirow{2}{*}{\makecell[c]{Backbone}} & Pre-training & \multicolumn{3}{c}{[avg(Org-var) / Org]\%$\downarrow$} \\
\cline{4-6}
&  & Datasets & N-Cal & Mini-ES & Mini-N \\
\midrule
MEM & ViT-B & N-Img$^+$ & 1.510 & 1.327 & 16.6$^*$ \\
ECDP & ViT-S & N-Img$^+$ & 1.969 & 2.863 & 14.0$^*$ \\
ECDP\& & ViT-S & EF-Img$^+$ & - & 2.549 & 4.1 \\
ECDDP & Swin-T & E-Tar & 2.963 & 2.437 & 4.7 \\
\rowcolor{gray!20}Ours & ViT-S & EF-Img$^+$ & 1.408 & 1.457 & \underline{2.7} \\
\rowcolor{gray!20}Ours & CViT-S & EF-Img$^+$ & \underline{0.769} & \underline{1.391} & 4.3 \\
\rowcolor{gray!20}Ours & Swin-T & EF-Img$^+$ & \textbf{0.447} & \textbf{1.321} & \textbf{0.7} \\
\bottomrule
\end{tabularx}
\begin{tablenotes}[para,flushleft]
`Org': performance on original test set.
`var': average performance on variant test sets.
\end{tablenotes}
\end{threeparttable}
\vspace{-2mm}
\end{table}

\begin{table}[t]
\scriptsize
\centering
\caption{Ablation study on different combination strategies.}
\vspace{-3mm}
\label{combination_abl}
\setlength\tabcolsep{0.14cm}
\belowrulesep=0pt
\aboverulesep=0pt
\renewcommand{\arraystretch}{0.8}
\begin{tabularx}{\columnwidth}{>
{\centering\arraybackslash}p{1.0cm}>
{\centering\arraybackslash}p{1.0cm}>
{\centering\arraybackslash}p{1.8cm}|>
{\centering\arraybackslash}p{1.5cm}>
{\centering\arraybackslash}p{1.75cm}}
\toprule
\multicolumn{3}{c|}{Combination Strategies} & \multicolumn{2}{c}{N-Caltech101 Top-1 Acc(\%)} \\
\midrule
MM & CL & Strategy & ViT & ConvViT \\
\midrule
\checkmark &  & - & 89.39 & 87.67 \\
 & \checkmark & - & 81.54 & 88.77 \\
\checkmark & \checkmark & MM-Trans-CL & \textbf{90.45} & \textbf{92.36} \\
\bottomrule
\end{tabularx}
\vspace{-2mm}
\end{table}

\begin{table}[t]
\scriptsize
\centering
\caption{Ablation study on different feature transition strategies.}
\vspace{-3mm}
\label{transition_abl}
\setlength\tabcolsep{0.14cm}
\belowrulesep=0pt
\aboverulesep=0pt
\renewcommand{\arraystretch}{0.8}
\begin{tabularx}{\columnwidth}{>
{\centering\arraybackslash}p{1.0cm}>
{\centering\arraybackslash}p{3.08cm}|>
{\centering\arraybackslash}p{1.5cm}>
{\centering\arraybackslash}p{1.75cm}}
\toprule
\multicolumn{2}{c|}{Feature Transition Strategies} & \multicolumn{2}{c}{N-Caltech101 Top-1 Acc(\%)} \\
\midrule
Trans & Object & ViT & ConvViT \\
\midrule
$\times$ & - & 89.93 & 92.01 \\
\checkmark & the whole model & 90.05 & 91.90 \\
\checkmark & only CL head & \textbf{90.45} & \textbf{92.36} \\
\bottomrule
\end{tabularx}
\vspace{-2mm}
\end{table}

\begin{table}[t]
\scriptsize
\centering
\caption{Ablation study on feature fusion strategies.}
\vspace{-3mm}
\label{fusion_abl}
\setlength\tabcolsep{0.14cm}
\belowrulesep=0pt
\aboverulesep=0pt
\renewcommand{\arraystretch}{0.8}
\begin{tabularx}{\columnwidth}{>
{\centering\arraybackslash}p{2.0cm}>
{\centering\arraybackslash}p{2.08cm}|>
{\centering\arraybackslash}p{1.5cm}>
{\centering\arraybackslash}p{1.75cm}}
\toprule
\multicolumn{2}{c|}{Feature Fusion Strategies} & \multicolumn{2}{c}{N-Caltech101 Top-1 Acc(\%)} \\
\midrule
Feature Fusion & Object & ViT & ConvViT \\
\midrule
$\times$ & H & 90.28 & 91.90 \\
\checkmark & H+L & \textbf{90.45} & \textbf{92.36} \\
\bottomrule
\end{tabularx}
\vspace{-2mm}
\end{table}

\subsection{Ablation study}\label{sec_abl}

\begin{figure}[t]
\centering
\includegraphics[width=0.7\columnwidth]{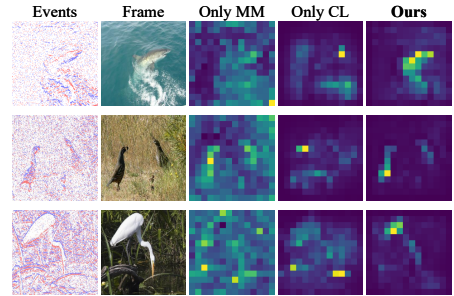}
\vspace{-3mm}
\caption{Attention map of different combination strategies.}
\vspace{-3mm}
\label{pr_attn_abl}
\end{figure}

\noindent\textbf{Combination strategy.}
As shown in Table~\ref{combination_abl} and Fig.~\ref{pr_attn_abl}, masked modeling (`MM') alone provides representational capability but results in lower performance due to semantic limitations and reduced discrimination, as it spreads attention across all regions with valid information.
Contrastive learning (`CL') performs better with strong guidance from high-level features, yet remains insufficient due to its susceptibility to being misled by denser but irrelevant events.
Preceding contrastive learning with masked modeling leads to significant improvement, demonstrating the effectiveness of our approach.

\begin{figure}[t]
\centering
\includegraphics[width=\columnwidth]{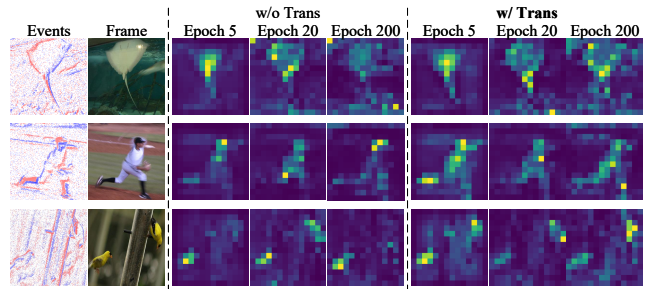}
\vspace{-7mm}
\caption{Attention shift with and without feature transition during last training stage.}
\vspace{-3mm}
\label{pr_attn_trans_abl}
\end{figure}

\begin{figure}[t]
\centering
\includegraphics[width=\columnwidth]{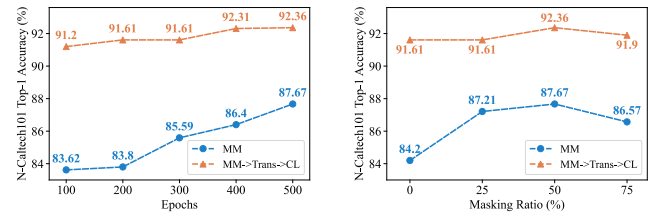}
\vspace{-7mm}
\caption{Different masking ratio and training time of masked modeling.}
\vspace{-3mm}
\label{abl_result}
\end{figure}

\noindent\textbf{Feature transition strategy.}
Table~\ref{transition_abl} shows that our feature transition (`Trans') method, which only updates the contrastive heads, outperforms methods that do not use transition or update the entire model, indicating that it is simple yet effective.
As shown in Fig.~\ref{pr_attn_trans_abl}, we visualize attention maps with and without feature transition at epochs 5, 20, and 200 (final) during the last training stage.
With `Trans', attention gradually shifts from broad coverage to more focused regions.
Without `Trans', attention may drift to irrelevant corners, partially cover the target, or miss objects entirely, highlighting its bridging role in transferring representations from masked modeling to contrastive learning.

\noindent\textbf{Feature fusion strategy.}
As shown in Table~\ref{fusion_abl}, the feature fusion module in the masked modeling stage improves performance, demonstrating its effectiveness in aiding edges and textures reconstruction and preserving high-level semantics from the upper layers, without adding notable computational cost.

\noindent\textbf{Masked modeling strategy.}
We experiment the masking ratio and training time for masked modeling using ConvViT backbone.
As shown in Fig.~\ref{abl_result}, the optimal masking ratio for our method is 50\%, which is lower than 75\% for MAE in images~\cite{he2022masked} and around 90\% by MAE for videos~\cite{tong2022videomae}, likely due to the sparsity of event data.
Longer training time improves the performance of our masked modeling, thereby enhancing the performance of the entire method.

\section{Conclusion}
\label{sec:conclusion}
We propose a novel and efficient event data pre-training framework that fully reveals latent information within events.
Our three-stage method progressively enables the model to extract enhanced information and improve semantic discrimination from sparse and noisy events, producing high-quality event representations.
Extensive experiments demonstrate superior performance on multiple downstream tasks and robustness across diverse sampling conditions.
Our framework offers a scalable, flexible solution for event-based learning, paving the way for future advances in the field.

\noindent \textbf{Acknowledgments}
This work is partially supported by National Natural Science Foundation of China under Grant No.62302041 and 62322204, and China National Postdoctoral Program under contract No.BX20230469.

\bibliographystyle{ACM-Reference-Format}
\bibliography{main}

\clearpage
\title{Supplementary Material}

\renewcommand{\shortauthors}{Zhu et al.}

\setcounter{page}{1}
\renewcommand{\thesection}{S\arabic{section}}
\setcounter{section}{0}
\renewcommand\thefigure{S\arabic{figure}}
\setcounter{figure}{0}
\renewcommand\thetable{S\arabic{table}}
\setcounter{table}{0}
\renewcommand\theequation{S\arabic{equation}}
\setcounter{equation}{0}

\section{Additional Discussion and Comparison with Other Methods}
\noindent\textbf{Recent advanced event representation methods.}
We discuss and compare several recent advanced event representation methods, including EvRepSL~\cite{qu2024evrepsl}, OpenESS~\cite{kong2024openess}, and EventBind~\cite{zhou2024eventbind}.
EvRepSL learns event representations at the original resolution by modeling the correlation between events and intensity, but only within regions where events occur.
In contrast, our method pre-trains a backbone capable of extracting fine-grained edge information and texture cues even in event-free regions, producing low-resolution yet high-dimensional feature representations.
Furthermore, it incorporates contrastive learning to improve semantic discrimination.
We compare our method with EvRepSL on four downstream task datasets: N-Caltech101~\cite{orchard2015converting}, DDD17~\cite{binas2017ddd17}, DSEC~\cite{gehrig2021dsec}, and MVSEC~\cite{zhu2018multivehicle}.
For N-Caltech101, the performance of EvRepSL is reported as in its original paper, where it uses a five-channel event representation as input, followed by three 1×1 convolutional layers and a global average pooling layer.
To ensure a fair comparison on semantic segmentation (DDD17, DSEC) and optical flow estimation (MVSEC), we implement EvRepSL using the same Transformer-based architecture as our method.
As shown in Table~\ref{evrepsl}, our method consistently outperforms EvRepSL, demonstrating the superiority of our pre-trained backbone, which can be fine-tuned more effectively than EvRepSL’s fixed event representations.
OpenESS and EventBind, on the other hand, are designed for fundamentally different settings.
Our self-supervised pre-training method does not rely on any labels and is fine-tuned on complete datasets for various downstream tasks.
In contrast, OpenESS and EventBind incorporate class-labeled text and adopt event-image-text contrastive learning to support open-set scene understanding, achieving strong performance in zero-shot and few-shot segmentation and object recognition.
They fall outside the scope of our setting and are therefore not included in our comparisons.

\noindent\textbf{Technical contributions of masked modeling.}
We are the first to propose a masked modeling approach that reconstructs masked temporal intensity difference maps from events.
Prior methods vary in their reconstruction designs and masking strategies, limiting their ability to align with event sampling principles and fully extract underlying edge and texture information from event data.
For example, Cho et al.~\cite{cho2023label} and LLM-EvRep~\cite{yu2025llm} reconstruct intensity images or misaligned edge maps, which do not precisely reflect the physical sampling properties of events.
MEM~\cite{klenk2024masked} attempts reconstruction on accumulated event images but remains vulnerable to noise and sparsity inherent in event data.
Meanwhile, EvRepSL restricts learning to regions where events occur, overlooking potentially valuable information present in event-free areas.

\noindent\textbf{Technical contributions of feature transition.}
Our feature transition method freezes the backbone weights during contrastive learning and only updates the randomly initialized contrastive head.
This prevents the degradation of features learned during masked modeling, which is important for representing noisy and sparse events.
The most similar prior method, Layer-Grafted Pre-training~\cite{jiang2023layer}, updates the backbone during contrastive learning using layer-wise decreasing learning rates after masked modeling, which still allows the randomly initialized contrastive head to degrade the learned features.
We conduct experiments replacing our feature transition method with layer-grafted pre-training.
Results in Table~\ref{layer_grafted} demonstrate the advantage of our approach.

\begin{table}[t]
\scriptsize
\centering
\caption{Comparison between our method and EvRepSL.}
\vspace{-3mm}
\label{evrepsl}
\setlength\tabcolsep{0.14cm}
\belowrulesep=0pt
\aboverulesep=0pt
\renewcommand{\arraystretch}{0.8}
\begin{threeparttable}
\begin{tabularx}{\columnwidth}{>
{\centering\arraybackslash}p{1.72cm}|>
{\centering\arraybackslash}p{1.5cm}|>
{\centering\arraybackslash}p{0.9cm}>
{\centering\arraybackslash}p{0.9cm}|>
{\centering\arraybackslash}p{2.0cm}}
\toprule
\multirow{2}{*}{Method} & Acc(\%)$\uparrow$ & \multicolumn{2}{c|}{mIOU(\%)$\uparrow$} & AEE$\downarrow$ \\
\cline{2-5}
 & N-Caltech101 & DDD17 & DSEC & MVSEC \\
\midrule
EvRepSL~\cite{yu2025llm} & 86.40 & 38.93 & 38.50 & 1.178/2.415/2.023 \\
\rowcolor{gray!20} Ours & \textbf{90.45} & \textbf{56.06} & \textbf{47.24} & \textbf{0.457/1.1076/0.768} \\
\bottomrule
\end{tabularx}
\begin{tablenotes}[para,flushleft]
`/’ separates the results on indoor\_flying1, indoor\_flying2, and indoor\_flying3 sequences in the MVSEC.
\end{tablenotes}
\end{threeparttable}
\vspace{-4mm}
\end{table}

\begin{table}[t]
\scriptsize
\centering
\caption{Comparison between our feature transition method and layer-grafted method.}
\vspace{-3mm}
\label{layer_grafted}
\setlength\tabcolsep{0.14cm}
\belowrulesep=0pt
\aboverulesep=0pt
\renewcommand{\arraystretch}{0.8}
\begin{threeparttable}
\begin{tabularx}{\columnwidth}{>
{\centering\arraybackslash}p{1.72cm}|>
{\centering\arraybackslash}p{1.5cm}|>
{\centering\arraybackslash}p{0.9cm}>
{\centering\arraybackslash}p{0.9cm}|>
{\centering\arraybackslash}p{2.0cm}}
\toprule
\multirow{2}{*}{Method} & Acc(\%)$\uparrow$ & \multicolumn{2}{c|}{mIOU(\%)$\uparrow$} & AEE$\downarrow$ \\
\cline{2-5}
 & N-Caltech101 & DDD17 & DSEC & MVSEC \\
\midrule
Layer-Grafted~\cite{jiang2023layer} & 88.89 & 54.95 & 47.14 & 0.488/1.138/0.844 \\
\rowcolor{gray!20} Trans(Ours) & \textbf{90.45} & \textbf{56.06} & \textbf{47.24} & \textbf{0.457/1.1076/0.768} \\
\bottomrule
\end{tabularx}
\begin{tablenotes}[para,flushleft]
`/’ separates the results on indoor\_flying1, indoor\_flying2, and indoor\_flying3 sequences in the MVSEC.
\end{tablenotes}
\end{threeparttable}
\vspace{-4mm}
\end{table}

\begin{table}[t]
\scriptsize
\centering
\caption{Comparison of the performance on sequence comprehension tasks.}
\vspace{-3mm}
\label{sequence_comprehension}
\setlength\tabcolsep{0.14cm}
\belowrulesep=0pt
\aboverulesep=0pt
\renewcommand{\arraystretch}{0.8}
\begin{threeparttable}
\begin{tabularx}{\columnwidth}{>
{\centering\arraybackslash}p{1.5cm}|>
{\centering\arraybackslash}p{0.9cm}|>
{\centering\arraybackslash}p{0.84cm}>
{\centering\arraybackslash}p{0.5cm}|>
{\centering\arraybackslash}p{1.5cm}>
{\centering\arraybackslash}p{1.5cm}}
\toprule
\multirow{2}{*}{Method} & \multirow{2}{*}{Backbone} & \multicolumn{2}{c|}{Pre-training} & \multicolumn{2}{c}{Top-1 Acc(\%)} \\
\cline{3-6}
 &  & Dataset & Time & DVS128 Gesture & UCF101-DVS \\
\midrule
\multicolumn{6}{l}{Task-specific event-based methods.} \\
\midrule
RG-CNN~\cite{bi2020graph} & - & - & - & 96.1 & 63.2 \\
TTPOINT~\cite{ren2023ttpoint} & - & - & - & 98.8 & 72.5 \\
EventMamba~\cite{ren2025rethinking} & - & - & - & 99.2 & 97.9 \\
\midrule
\multicolumn{4}{l}{Backbone training from scratch.} \\
\midrule
Ours & ViT-S & - & - & 80.5 & 83.3 \\
Ours & CViT-S & - & - & 87.5 & 88.1 \\
Ours & Swin-T & - & - & 87.5 & 86.5 \\
\midrule
\multicolumn{6}{l}{Event-based self-supervised pre-training methods.} \\
\midrule
MEM~\cite{klenk2024masked} & ViT-B & N-Img & 275 & \textbf{97.3} & 92.1 \\
ECDP~\cite{yang2023event} & ViT-S & N-Img$^+$ & 267 & 96.9 & 92.1 \\
ECDDP~\cite{yang2024event} & Swin-T & E-Tar & 152 & 96.1 & 92.4 \\
\rowcolor{gray!20} Ours & ViT-S & EF-Img$^+$ & 80 & \textbf{97.3} & 92.6 \\
\rowcolor{gray!20} Ours & CViT-S & EF-Img$^+$ & 88 & \textbf{97.3} & \textbf{93.3} \\
\rowcolor{gray!20} Ours & Swin-T & EF-Img$^+$ & 91 & 96.1 & \textbf{93.3} \\
\bottomrule
\end{tabularx}
\begin{tablenotes}[para,flushleft]
Time: the measured pre-training time.
`+': method using paired RGB images with events.
\end{tablenotes}
\end{threeparttable}
\vspace{-4mm}
\end{table}

\begin{table*}[t]
\scriptsize
\centering
\caption{Comprehensive evaluation of pre-training cost.}
\vspace{-3mm}
\label{training_cost}
\setlength\tabcolsep{0.14cm}
\belowrulesep=0pt
\aboverulesep=0pt
\renewcommand{\arraystretch}{0.8}
\begin{threeparttable}
\begin{tabularx}{\linewidth}{>
{\centering\arraybackslash}p{2.07cm}|>
{\centering\arraybackslash}p{2.0cm}|>
{\centering\arraybackslash}p{3.0cm}>
{\centering\arraybackslash}p{3.0cm}>
{\centering\arraybackslash}p{3.0cm}>
{\centering\arraybackslash}p{3.0cm}
}
\toprule
Method & Backbone & Params(M) & GFLOPs(G) & GPU memory(MB) & Training time(h) \\
\midrule
MEM & ViT-B & 16.67/111.6 & 79.09/9.68 & 17277/8598 & 150/1215=275 \\
ECDP & ViT-S & 64.74 & 3.19 & 3526 & 266.7 \\
ECDDP & Swin-T & 102.95 & 35.38 & 20420 & 151.7 \\
\rowcolor{gray!20} Ours & ViT-S & 28.27/23.29/45.08 & 3.45/8.85/8.85 & 4214/2838/7004 & 62.5/0.6/16.7=\textbf{79.8} \\
\rowcolor{gray!20} Ours & CViT-S & 29.53/23.29/45.16 & 5.76/10.75/10.75 & 7264/2870/9461 & 	65.3/0.75/21.7=\underline{87.75} \\
\rowcolor{gray!20} Ours & Swin-T & 42.56/30.96/58.48 & 2.97/5.91/5.91 & 4587/3062/7375 & 62.5/0.83/27.8=91.13 \\
\bottomrule
\end{tabularx}
\begin{tablenotes}[para,flushleft]
`/' separates the results from different pre-training stages.
\end{tablenotes}
\end{threeparttable}
\vspace{-4mm}
\end{table*}

\begin{table*}[t]
\scriptsize
\centering
\caption{Evaluation of model complexity and inference time on downstream tasks.}
\vspace{-3mm}
\label{downstream_cost}
\setlength\tabcolsep{0.14cm}
\belowrulesep=0pt
\aboverulesep=0pt
\renewcommand{\arraystretch}{0.8}
\begin{threeparttable}
\begin{tabularx}{\linewidth}{>
{\centering\arraybackslash}p{2.07cm}|>
{\centering\arraybackslash}p{2.0cm}|>
{\centering\arraybackslash}p{3.0cm}>
{\centering\arraybackslash}p{3.0cm}>
{\centering\arraybackslash}p{3.0cm}>
{\centering\arraybackslash}p{3.0cm}
}
\toprule
Method & Backbone & Params(M) & GFLOPs(G) & GPU memory(MB) & Inference time(ms) \\
\midrule
MEM & ViT-B & 85.85/106.88/106.87 & 12.69/16.62/16.62 & 	8005(1110)/12540(1835)/12680(1548) & 21.25/26.47/31.89 \\
ECDP & ViT-S & 21.57/39.35/39.35 & 4.26/7.66/7.66 & 4634(555)/8691(1086)/8808(1205) & 19.34/23.22/22.32 \\
ECDDP & Swin-T & 27.6/47.12/47.12 & 4.4/27.37/27.37 & 5451(1305)/11240(4310)/11400(4516) & 37.24/28.53/32.67 \\
\rowcolor{gray!20} Ours & ViT-S & 21.83/39.65/39.65 & 4.28/7.68/7.68 & 4704(681)/8722(1232)/8906(1416) & 20.2/23.42/24.01 \\
\rowcolor{gray!20} Ours & CViT-S & 21.91/40.91/40.91 & 6.18/29.11/29.1 & 7330(1121)/13927(3849)/14110(4035) & 	20.81/24.46/26.17 \\
\rowcolor{gray!20} Ours & Swin-T & 27.6/52.43/52.42 & 4.39/27.37/29.1 & 5090(1201)/11045(4227)/11230(4413) & 178.32/141.74/161.63 \\
\bottomrule
\end{tabularx}
\begin{tablenotes}[para,flushleft]
`/' separates the results on N-Caltech101, DDD17 and MVSEC.
GPU memory values outside brackets is for training and inside for testing.
\end{tablenotes}
\end{threeparttable}
\vspace{-4mm}
\end{table*}

\begin{table}[t]
\scriptsize
\centering
\caption{Ablation study on different combination strategies.}
\vspace{-3mm}
\label{combination_abl}
\setlength\tabcolsep{0.14cm}
\belowrulesep=0pt
\aboverulesep=0pt
\renewcommand{\arraystretch}{0.8}
\begin{tabularx}{\columnwidth}{>
{\centering\arraybackslash}p{1.0cm}>
{\centering\arraybackslash}p{1.0cm}>
{\centering\arraybackslash}p{1.8cm}|>
{\centering\arraybackslash}p{1.5cm}|>
{\centering\arraybackslash}p{1.73cm}}
\toprule
\multicolumn{3}{c|}{Combination Strategies} & mIOU(\%)$\uparrow$ & AEE$\downarrow$ \\
\midrule
MM & CL & Strategy & DDD17 & MVSEC \\
\midrule
\checkmark &  & - & 54.97 & 0.486/1.152/0.827 \\
 & \checkmark & - & 50.04 & 0.575/1.369/0.994 \\
\checkmark & \checkmark & MM-Trans-CL & \textbf{56.06} & \textbf{0.457/1.076/0.768} \\
\bottomrule
\end{tabularx}
\vspace{-4mm}
\end{table}

\begin{table}[t]
\scriptsize
\centering
\caption{Ablation study on different feature transition strategies.}
\vspace{-3mm}
\label{transition_abl}
\setlength\tabcolsep{0.14cm}
\belowrulesep=0pt
\aboverulesep=0pt
\renewcommand{\arraystretch}{0.8}
\begin{tabularx}{\columnwidth}{>
{\centering\arraybackslash}p{1.0cm}>
{\centering\arraybackslash}p{3.06cm}|>
{\centering\arraybackslash}p{1.5cm}|>
{\centering\arraybackslash}p{1.75cm}
}
\toprule
\multicolumn{2}{c|}{Feature Transition Strategies} & mIOU(\%)$\uparrow$ & AEE$\downarrow$ \\
\midrule
Trans & Object & DDD17 & MVSEC \\
\midrule
$\times$ & - & 55.65 & \textbf{0.457}/1.097/0.795 \\
\checkmark & the whole model & 55.84 & 0.462/1.090/0.780 \\
\checkmark & only CL head & \textbf{56.06} & \textbf{0.457/1.076/0.768} \\
\bottomrule
\end{tabularx}
\vspace{-4mm}
\end{table}

\begin{table}[t]
\scriptsize
\centering
\caption{Ablation study on feature fusion strategies.}
\vspace{-3mm}
\label{fusion_abl}
\setlength\tabcolsep{0.14cm}
\belowrulesep=0pt
\aboverulesep=0pt
\renewcommand{\arraystretch}{0.8}
\begin{tabularx}{\columnwidth}{>
{\centering\arraybackslash}p{2.0cm}>
{\centering\arraybackslash}p{2.06cm}|>
{\centering\arraybackslash}p{1.5cm}|>
{\centering\arraybackslash}p{1.75cm}}
\toprule
\multicolumn{2}{c|}{Feature Fusion Strategies} & mIOU(\%)$\uparrow$ & AEE$\downarrow$ \\
\midrule
Feature Fusion & Object & DDD17 & MVSEC \\
\midrule
$\times$ & H & 55.74 & 0.460/1.107/0.808 \\
\checkmark & H+L & \textbf{56.06} & \textbf{0.457/1.076/0.768} \\
\bottomrule
\end{tabularx}
\vspace{-4mm}
\end{table}

\begin{table}[t]
\scriptsize
\centering
\caption{Ablation study on different masking strategies on dense prediction tasks.}
\vspace{-3mm}
\label{mask_abl}
\setlength\tabcolsep{0.14cm}
\belowrulesep=0pt
\aboverulesep=0pt
\renewcommand{\arraystretch}{0.8}
\begin{tabularx}{\columnwidth}{>
{\centering\arraybackslash}p{2.3cm}|>
{\centering\arraybackslash}p{1.5cm}|>
{\centering\arraybackslash}p{1.5cm}|>
{\centering\arraybackslash}p{2.0cm}
}
\toprule
\multirow{2}{*}{Masking Strategies} & Acc(\%)$\uparrow$ & mIOU(\%)$\uparrow$ & AEE$\downarrow$ \\
\cline{2-4}
 & N-Caltech101 & DDD17 & MVSEC \\
\midrule
anti-density & 91.03 & 58.85 & 0.452/1.080/0.783 \\
density & 90.10 & 58.58 & 0.453/1.118/0.823 \\
random & \textbf{92.36} & \textbf{60.53} & \textbf{0.434/1.035/0.778} \\
\bottomrule
\end{tabularx}
\vspace{-4mm}
\end{table}

\noindent\textbf{Evaluation on dynamic sequence tasks.}
Our method, similar to MEM, ECDP~\cite{yang2023event}, and ECDDP~\cite{yang2024event}, is not specifically designed for action recognition tasks that emphasize temporal clues.
We do not employ specialized architectures such as recurrent temporal structures.
Instead, we use event voxel representations to retain temporal information in a general and lightweight manner.
Performance comparisons on DVS128 Gesture~\cite{amir2017low} and UCF101-DVS~\cite{bi2020graph} are presented in Table~\ref{sequence_comprehension}.
While some task-specific methods outperform generic pre-training approaches due to their tailored designs, our method generally achieves better performance than other pre-training methods—particularly on UCF101-DVS, which involves more diverse categories.

\section{Evaluation of Training and Inference Cost}

\noindent\textbf{Comprehensive evaluation of pre-training cost.}
We evaluate the pre-training cost—including model parameters, computation cost (GFLOPs), GPU memory usage, and pre-training time—for all methods on a single NVIDIA GeForce RTX 3090 GPU with a batch size of 64, separately considering different pre-training stages.
As shown in Table~\ref{training_cost}, our method has low parameters, computation cost, and GPU memory usage, and especially requires shorter training time compared to existing pre-training methods.
This highlights its efficiency in achieving higher performance with lower training cost.

\noindent\textbf{Evaluation of model complexity and inference time on downstream tasks.}
Table~\ref{downstream_cost} shows the evaluation of model complexity and inference time for downstream tasks under the same setup as the pre-training cost evaluation.
In general, our method has low parameter count, computation cost, and GPU memory usage.
Our ViT-based and ConvViT-based methods achieve short inference times, while the Swin-based methods take relatively longer due to the optimal grouping strategy employed in the MAE-compatible Swin Transformer~\cite{huang2022green}.

\begin{figure}[t]
\centering
\includegraphics[width=\columnwidth]{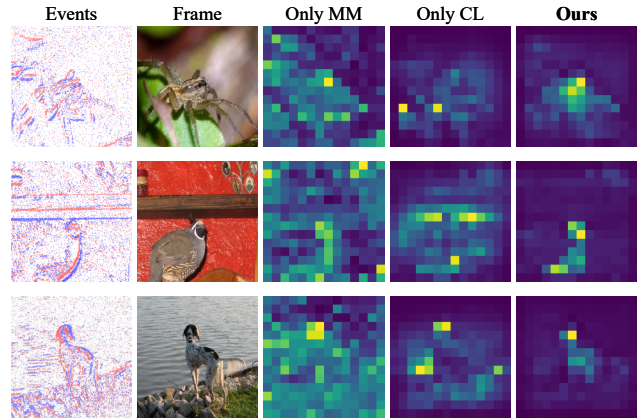}
\vspace{-3mm}
\caption{
\textbf{More attention map comparison of different combination
strategies.}}
\vspace{-3mm}
\label{pr_attn_abl_supp}
\end{figure}

\section{Additional Ablation Study}
\noindent\textbf{Ablation study on dense prediction tasks.}
We experiment with different combination, feature transition, and feature fusion strategies on dense prediction tasks using ViT backbone in Tables~\ref{combination_abl},~\ref{transition_abl}, and~\ref{fusion_abl}, respectively (corresponding to Tables 6, 7, and 8 in the main text).
The observed trends are consistent with those in global recognition tasks.

\noindent\textbf{Masking strategy.}
We experiment with different masking strategies for masked modeling using ConvViT backbone in Table~\ref{mask_abl}.
Random masking during the masked modeling stage outperforms strategies that mask denser patches, as in ECDP~\cite{yang2023event}, or sparser patches, demonstrating the effectiveness of preserving both dense and sparse regions for more comprehensive representation.

\noindent\textbf{Additional results of different combination strategies.}
Additional attention map comparison results can be seen in Fig.~\ref{pr_attn_abl_supp}.

\section{EF-ImageNet}

\noindent\textbf{Generation details.}
We simulate our event-frame alignment pre-training
dataset, EF-ImageNet, using the top 10\% of each class based on image serial numbers in
ImageNet-1K~\cite{deng2009imagenet}.
The generation process includes two steps:

\textbf{i) Image to video.}
We first resize the RGB images into $280 \times 280$ and set a cropping area of $224 \times 224$ in the center.
Then, we generate two sets of motion parameters, each including four common modes: translation, scaling, rotation, and perspective.
The first set is used to randomly reposition the original images, which are then used as the starting frames of the video.
This process aims to enhance the diversity of our dataset.
The second set is used to move the start frame to the end frame, in order to obtain the whole video.
We set boundary restrictions to ensure the cropping area stays within the images during movement, establish operation thresholds to retain as much original content as possible in the video, and restrict the motion modes of video frames to include translation, making it conform to reality.
The proportions of each motion mode within the two sets of parameters are listed in Table~\ref{motion}.
Finally, based on these parameters, we obtain 12 video frames for each image by moving them accordingly, apply interpolation to minimize blurring, and then synthesize the video at a frame rate of 30 frames per second (fps).
The motion parameters and the position of each frame are recorded in files.

\textbf{ii) Video to events.}
We use v2e~\cite{hu2021v2e} to generate events from videos, configuring it based on its `noisy' model.
The only adjustment is changing the `cutoff\_hz' parameter from 30 to 15 to balance timestamp resolution and the cutoff frequency of the DVS photoreceptor.
Details are provided in Table~\ref{v2e_params}.
Besides, we exclude the first inter-frame event sequence due to its notable sparsity.
Therefore, we finally obtain 11 frames aligned with 10 inter-frame event sequences for each image.

\begin{table}[t]
\scriptsize
\centering
\caption{The proportions of each motion mode within the two sets of parameters of our EF-ImageNet.}
\vspace{-3mm}
\label{motion}
\setlength\tabcolsep{0.1cm}
\belowrulesep=0pt
\aboverulesep=0pt
\begin{threeparttable}
\begin{tabularx}{\columnwidth}{>
{\centering\arraybackslash}p{1.65cm}>
{\centering\arraybackslash}p{2.0cm}|>
{\centering\arraybackslash}p{2.0cm}>
{\centering\arraybackslash}p{2.0cm}}
\toprule
\multicolumn{2}{c|}{\multirow{2}{*}{Mode}} & \multicolumn{2}{c}{Proportion} \\
\cline{3-4}
 &  & org to start(\%) & start to end(\%) \\
\midrule
\multirow{9}{*}{translation} & left & 15.93 & 20.60 \\
 & right & 15.81 & 20.39 \\
 & up & 16.13 & 20.53 \\
 & down & 15.67 & 20.86 \\
 & upper left & 3.69 & 4.47 \\
 & upper right & 3.76 & 4.30 \\
 & lower left & 3.74 & 4.40 \\
 & lower right & 3.77 & 4.46 \\
 & no translation & 21.49 & 0.00 \\
\midrule
\multirow{3}{*}{scaling} & zoom in & 22.54 & 20.54 \\
 & zoom out & 9.99 & 12.64 \\
 & no scaling & 67.48 & 66.82 \\
\midrule
\multirow{3}{*}{rotation} & clockwise & 11.60 & 13.11 \\
 & anticlockwise & 11.65 & 13.04 \\
 & no rotation & 76.75 & 73.85 \\
\midrule
\multirow{2}{*}{perspective} & perspective & 34.72 & 36.60 \\
 & no perspective & 65.28 & 63.40 \\
\bottomrule
\end{tabularx}
\begin{tablenotes}[para,flushleft]
\end{tablenotes}
\end{threeparttable}
\vspace{-4mm}
\end{table}

\begin{table}[t]
\scriptsize
\centering
\caption{Configuration of v2e~\cite{hu2021v2e} to generate events from videos.}
\vspace{-3mm}
\label{v2e_params}
\setlength\tabcolsep{0.1cm}
\belowrulesep=0pt
\aboverulesep=0pt
\begin{threeparttable}
\begin{tabularx}{\columnwidth}{>
{\centering\arraybackslash}p{4.05cm}|>
{\centering\arraybackslash}p{4.0cm}}
\toprule
Parameters & Value \\
\midrule
timestamp\_resolution & 0.003 \\
auto\_timestamp\_resolution & False \\
dvs\_exposure & duration 0.005 \\
pos\_thres & 0.2 \\
neg\_thres & 0.2 \\
sigma\_thres & 0.05 \\
cutoff\_hz & 15 \\
leak\_rate\_hz & 0.1 \\
leak\_jitter\_fraction & 0.1 \\
noise\_rate\_cov\_decades & 0.1 \\
shot\_noise\_rate\_hz & 5.0 \\
\bottomrule
\end{tabularx}
\begin{tablenotes}[para,flushleft]
\end{tablenotes}
\end{threeparttable}
\vspace{-4mm}
\end{table}

\noindent\textbf{Discussion and future work.}
Currently, there is no large-scale real-world dataset that provides temporally and spatially aligned event streams and video frames suitable for masked modeling.
To address this, we simulate event data to construct the pre-training dataset.
Real-world collection requires physically moving an event camera over displayed images, which involves mechanical devices, restricts motion diversity, and introduces display-induced noise.
In contrast, our simulation-based approach is more cost-effective, allows for diverse motion patterns, and avoids external noise interference.
Despite being trained on synthetic data, our method generalizes well and achieves strong performance in various real-world scenarios.
We are working toward building corresponding real-world datasets and exploring improved pre-training strategies to further enhance event representation.

\section{More explanation of pre-training results}
\begin{figure*}[t]
\centering
\includegraphics[width=0.8\textwidth]{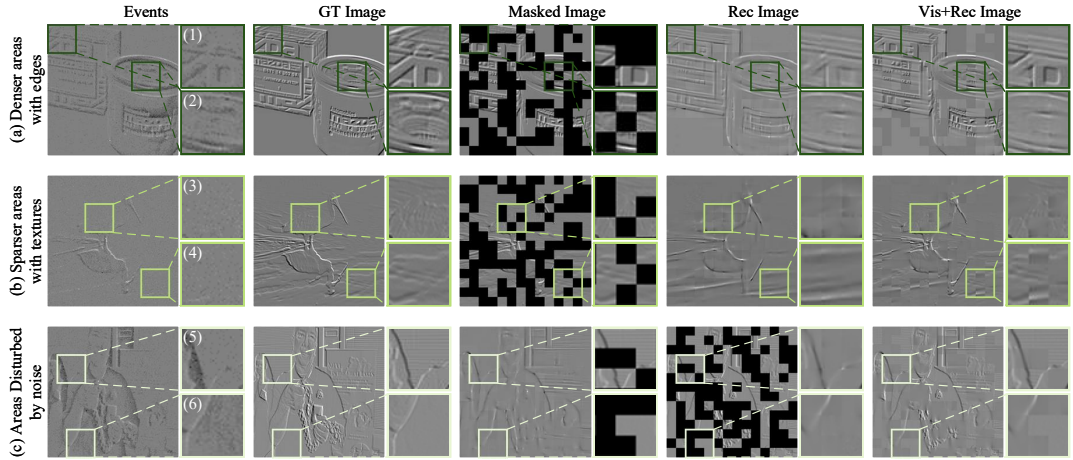}
\vspace{-3mm}
\caption{
\textbf{More masked modeling results. Illustration of (a)-(c) and (1)-(6) are the same as those in Fig. 4.}}
\label{pr_rec_supp}
\end{figure*}

\noindent\textbf{Additional masked modeling results.}
The masked modeling results are obtained from our ConvViT-based model.
Additional masked modeling results are shown in Fig.~\ref{pr_rec_supp}.

\begin{figure*}[t]
\centering
\includegraphics[width=0.8\textwidth]{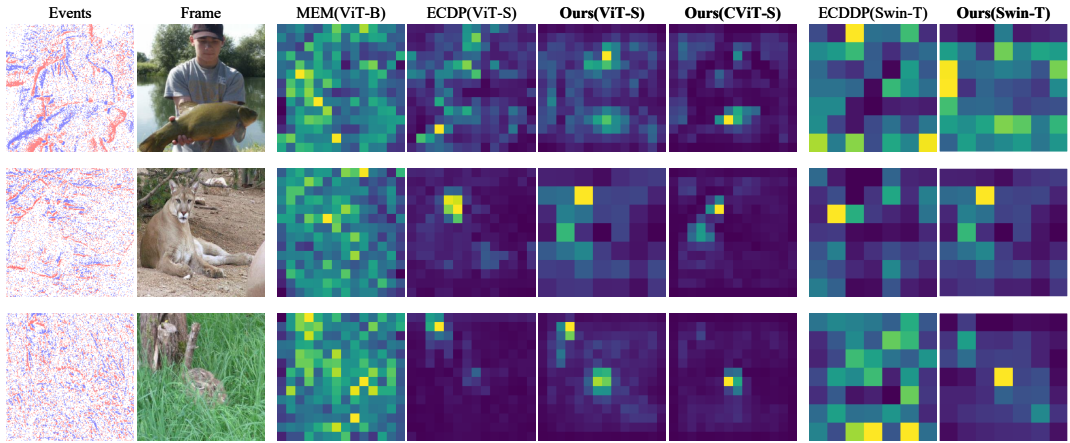}
\vspace{-3mm}
\caption{
\textbf{More attention map comparison of different methods.}}
\label{pr_attn_supp}
\end{figure*}

\noindent\textbf{Calculation methods of attention maps and additional results.}
MEM~\cite{klenk2024masked}, ECDDP~\cite{yang2024event}, and our method use features from all tokens to represent event features, so attention maps are obtained by averaging attention weights across all blocks.
This averaging can produce smaller but more widespread attention values in areas with stronger local relationships.
In contrast, ECDP~\cite{yang2023event} relies only on the cls token, deriving its attention map solely from that token’s attention results.
Additional attention map results are shown in Fig.~\ref{pr_attn_supp}.

\section{Full comparison results on downstream tasks}
\noindent\textbf{Full comparison results on global recognition tasks.}
Results of the mini N-ImageNet across all 10 test sets (original and variants with 5 motion trajectories and 4 lighting conditions) are detailed in Table~\ref{mini_img}.

\begin{figure*}[t]
\centering
\includegraphics[width=\textwidth]{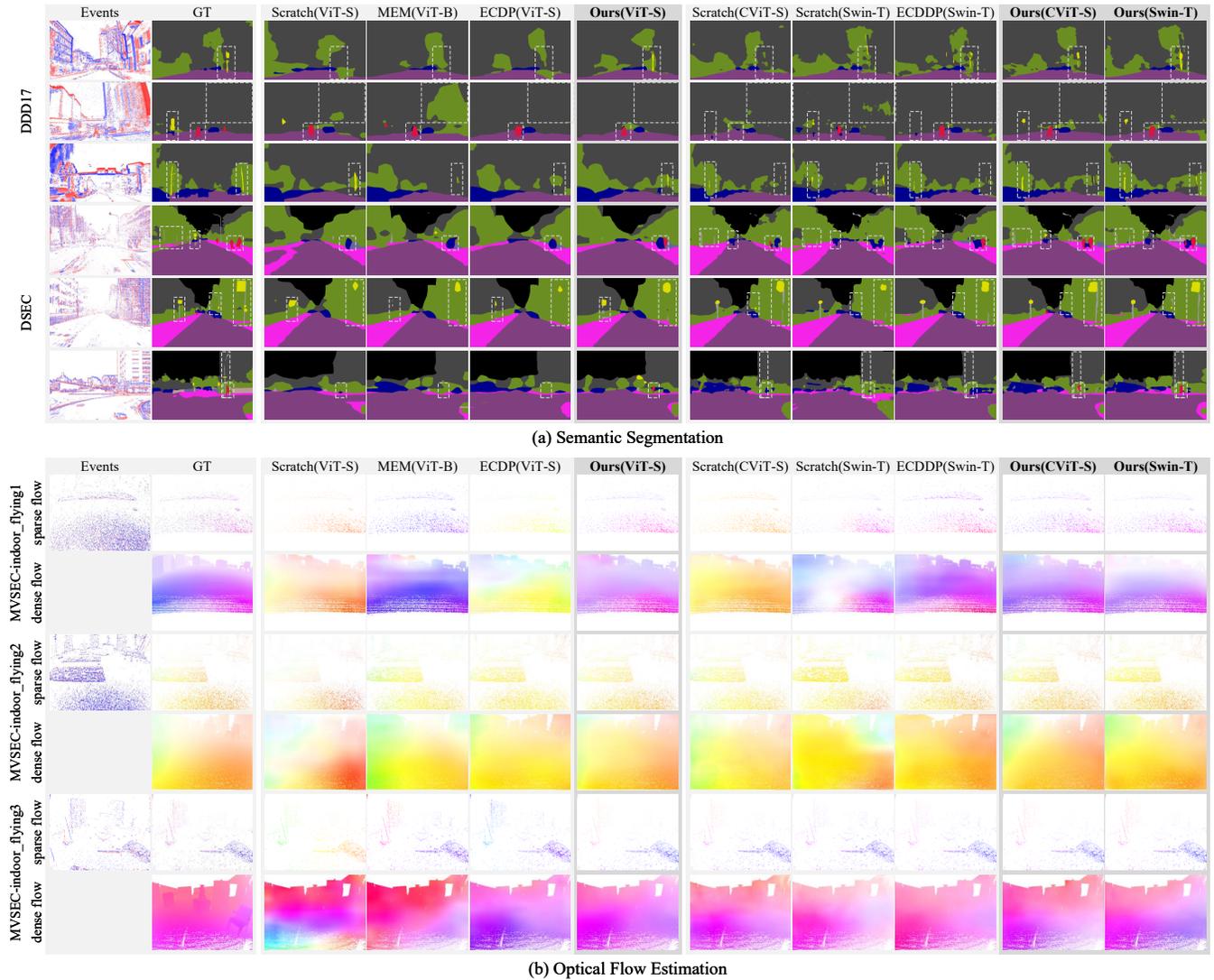}
\vspace{-7mm}
\caption{More qualitative results on semantic segmentation and optical flow estimation.}
\label{ft_results_supp}
\end{figure*}

\noindent\textbf{Additional comparison results on dense prediction tasks.}
The complete comparison results for the semantic segmentation and optical flow estimation tasks are shown in Table~\ref{sem_seg_supp} and~\ref{flow_supp}, respectively.
Additional qualitative results are shown in Fig.~\ref{ft_results_supp}.

\section{More experimental details}
\noindent\textbf{Implementation details of downstream tasks.}
For global recognition tasks (including object recognition), we simply append a linear classification head after the backbone network.
We split N-Caltech101~\cite{orchard2015converting} and CIFAR10-DVS~\cite{li2017cifar10} into training and testing sets following the same protocol used in ECDP~\cite{yang2023event}, MEM~\cite{klenk2024masked}, and Huang et al.~\cite{huang2024data} to ensure a fair comparison.
For dense prediction tasks (including semantic segmentation and optical flow estimation), we append UperNet~\cite{xiao2018unified} as the decoder head and FCN~\cite{long2015fully} as an auxiliary head, following the setup in BEIT~\cite{bao2021beit}.
We adopt the same train-test splits of MVSEC~\cite{zhu2018multivehicle} as used in ECDP to ensure consistency.

\noindent\textbf{Implementation details of robustness study.}
The details of three variant test sets are as follows:
i) Sparse: reduce the fixed number of events by $\frac{1}{5}$.
ii) Noise: randomly erase and add events, with the total number of modifications ranging from 0.001 to 0.01 times the original number of events.
iii) Sparse + Noise: a combination of both.
The complete experimental results are shown in Table~\ref{robust_small}.

\noindent\textbf{Training hyperparameters.}
We outline the training hyperparameters as follows: Table~\ref{pr_params} contains the hyperparameters for the three-stage pre-training process, Table~\ref{obj_params} details the hyperparameters for the fine-tuning on object recognition tasks, and Table~\ref{seg_flow_params} lists the hyperparameters for the fine-tuning on semantic segmentation, optical flow estimation and action recognition tasks.

\noindent\textbf{Details of retraining ECDP.}
We retrain ECDP using the same pre-training datasets as ours.
For a fairer comparison, we adjust some training details to align with our setup, including using the pre-trained ViT-B/16 CLIP~\cite{radford2021learning} network for RGB image feature extraction, implementing a queue of length 1024 for negative sample storage, and adopting data augmentation strategies similar to ours while keeping ECDP’s other original parameters unchanged.

\begin{table*}[t]
\scriptsize
\caption{Full experimental results on object recognition tasks using Mini N-ImageNet.}
\vspace{-3mm}
\label{mini_img}
\setlength\tabcolsep{0.1cm}
\belowrulesep=0pt
\aboverulesep=0pt
\begin{threeparttable}
\resizebox{\textwidth}{!}{
\begin{tabular*}{\linewidth}{>
{\centering\arraybackslash}p{1.1cm}|>
{\centering\arraybackslash}p{0.9cm}|>
{\centering\arraybackslash}p{0.8cm}>
{\centering\arraybackslash}p{0.8cm}|>
{\centering\arraybackslash}p{0.78cm}>
{\centering\arraybackslash}p{0.78cm}>
{\centering\arraybackslash}p{0.78cm}>
{\centering\arraybackslash}p{0.78cm}>
{\centering\arraybackslash}p{0.78cm}>
{\centering\arraybackslash}p{0.78cm}>
{\centering\arraybackslash}p{0.78cm}>
{\centering\arraybackslash}p{0.78cm}>
{\centering\arraybackslash}p{0.78cm}>
{\centering\arraybackslash}p{0.78cm}|>
{\centering\arraybackslash}p{0.78cm}>
{\centering\arraybackslash}p{0.78cm}>
{\centering\arraybackslash}p{1.36cm}}
\toprule
\multirow{3}{*}{Method} & \multirow{3}{*}{Backbone} & \multicolumn{2}{c|}{Pre-training} & \multicolumn{13}{c}{Mini N-ImageNet Top-1 Acc(\%)} \\
\cline{3-17}
& & Dataset & Time & Org & mode\_1 & mode\_3 & mode\_5 & mode\_6 & mode\_7 & bright\_4 & bright\_5 & bright\_6 & bright\_7 & avg(all)$\uparrow$ & avg(var)$\uparrow$ & \makecell[c]{[avg(Org-var)\\ / Org]\%$\downarrow$} \\
\midrule
\multicolumn{17}{l}{Backbone training from scratch.} \\
\midrule
Ours & ViT-S & - & - & 33.95 & 32.47 & 29.03 & 31.21 & 36.06 & 25.56 & 18.47 & 25.60 & 33.45 & 32.39 & 29.82 & 29.36 & 13.5 \\  
Ours & CViT-S & - & - & 45.53 & 44.31 & 41.10 & 37.38 & 41.51 & 32.31 & 21.90 & 28.95 & 39.02 & 37.64 & 36.90 & 36.01 & 20.9 \\  
Ours & Swin-T & - & - & 47.64 & 44.95 & 41.91 & 42.65 & 49.76 & 36.24 & 24.80 & 34.90 & 45.49 & 43.39 & 41.17 & 40.45 & 15.1 \\  
\midrule
\multicolumn{17}{l}{Event-based self-supervised pre-training methods.} \\
\midrule
MEM~\cite{klenk2024masked} & ViT-B & N-Img & 275 & 65.19$^*$ & 62.78$^*$ & 59.90$^*$ & 55.23$^*$ & 61.20$^*$ & 51.44$^*$ & 36.00$^*$ & 47.28$^*$ & 58.03$^*$ & 56.81$^*$ & 55.44$^*$ & 54.36$^*$ & 16.6$^*$ \\  
ECDP~\cite{yang2023event} & ViT-S & N-Img$^+$ & 267 & 72.35$^*$ & 71.37$^*$ & 67.41$^*$ & 63.42$^*$ & 69.21$^*$ & 57.33$^*$ & 45.65$^*$ & 56.73$^*$ & 65.34$^*$ & 63.64$^*$ & 63.25$^*$ & 62.23$^*$ & 14.0$^*$ \\  
ECDP\$~\cite{yang2023event} & ViT-S & EF-Img$^+$ & 63 & 41.81 & 39.58 & 36.30 & 43.45 & 49.26 & 37.46 & 27.14 & 37.18 & 46.09 & 44.35 & 40.26 & 40.09 & 4.1 \\
ECDDP~\cite{yang2024event} & Swin-T & E-Tar & 152 & 46.50 & 43.71 & 39.92 & 47.98 & 54.27 & 42.39 & 29.03 & 39.66 & 51.08 & 48.80 & 44.33 & 44.09 & 4.7 \\  
\rowcolor{gray!20} Ours & ViT-S & EF-Img$^+$ & 97 & 46.67 & 42.87 & 40.48 & 51.94 & 59.03 & 47.92 & 34.11 & 46.27 & 55.97 & 52.92 & 47.82 & 47.95 & 2.7 \\  
\rowcolor{gray!20} Ours & CViT-S & EF-Img$^+$ & 88 & 58.07 & 54.79 & 53.29 & 58.49 & 65.56 & 54.45 & 39.70 & 50.48 & 62.34 & 60.78 & 55.80 & 55.55 & 4.3 \\  
\rowcolor{gray!20} Ours & Swin-T & EF-Img$^+$ & 91 & 51.06 & 48.24 & 47.02 & 54.05 & 61.20 & 50.64 & 34.72 & 46.23 & 57.99 & 56.31 & 50.75 & 50.71 & 0.7 \\  
\rowcolor{gray!20} \makecell[c]{Ours} & ViT-S & EF+N$^+$ & 97 & 50.26$^{(*)}$ & 46.41$^{(*)}$ & 43.21$^{(*)}$ & 59.68$^{(*)}$ & 63.90$^{(*)}$ & 52.56$^{(*)}$ & 39.42$^{(*)}$ & 51.10$^{(*)}$ & 61.44$^{(*)}$ & 58.85$^{(*)}$ & 52.68$^{(*)}$ & 52.95$^{(*)}$ & 5.4$^{(*)}$ \\  
\rowcolor{gray!20} \makecell[c]{Ours} & CViT-S & EF+N$^+$ & 109 & 60.08$^{(*)}$ & 57.27$^{(*)}$ & 54.19$^{(*)}$ & 65.77$^{(*)}$ & 71.03$^{(*)}$ & 60.54$^{(*)}$ & 48.80$^{(*)}$ & 59.92$^{(*)}$ & 68.73$^{(*)}$ & 67.51$^{(*)}$ & 61.38$^{(*)}$ & 61.53$^{(*)}$ & 2.4$^{(*)}$ \\
\bottomrule
\end{tabular*}}
\begin{tablenotes}[para,flushleft]
`+': method using paired RGB images with events.
`*': method fine-tuning on the same dataset as pre-training, resulting in higher performance on this dataset.
`(*)': method using partly overlapping datasets for pre-training and fine-tuning.
`\&': method retrained with the same pre-training datasets as ours.
\end{tablenotes}
\end{threeparttable}
\vspace{-1mm}
\end{table*}

\begin{table*}[t]
\scriptsize
\centering
\caption{Full comparison of the performance on semantic segmentation tasks.}
\vspace{-3mm}
\label{sem_seg_supp}
\setlength\tabcolsep{0.1cm}
\belowrulesep=0pt
\aboverulesep=0pt
\begin{threeparttable}
\begin{tabularx}{\linewidth}{>
{\centering\arraybackslash}p{3.0cm}|>
{\centering\arraybackslash}p{1.8cm}>
{\centering\arraybackslash}p{1.0cm}|>
{\centering\arraybackslash}p{1.8cm}>
{\centering\arraybackslash}p{0.8cm}>
{\centering\arraybackslash}p{1.0cm}|>
{\centering\arraybackslash}p{1.58cm}>
{\centering\arraybackslash}p{1.58cm}|>
{\centering\arraybackslash}p{1.58cm}>
{\centering\arraybackslash}p{1.58cm}}
\toprule  
\multirow{2}{*}{Method} & \multirow{2}{*}{Backbone} & \multirow{2}{*}{Params} & \multicolumn{3}{c|}{Pre-training} & \multicolumn{2}{c|}{DDD17} & \multicolumn{2}{c}{DSEC} \\
\cline{4-10}
 &  &  & Dataset & Epochs & Time & mIOU(\%) & mACC(\%) & mIOU(\%) & mACC(\%) \\
\midrule  
\multicolumn{10}{l}{Task-specific event-based methods.} \\
\midrule
EV-SegNet~\cite{alonso2019ev} & - & - & - & - & - & 54.81 & - & 51.76 & - \\
ESS~\cite{sun2022ess} & - & - & - & - & - & \textbf{61.37}$^+$ & - & \textbf{53.29}$^+$ & - \\
\midrule  
\multicolumn{10}{l}{Backbone training from scratch.} \\
\midrule
MEM~\cite{klenk2024masked} & ViT-B/16 & 85.8M & - & - & - & - & - & 42.79 & 49.90 \\
ECDP~\cite{yang2023event} & ViT-S/16 & 21.6M & - & - & - & 48.76 & - & 40.53 & - \\
Ours & ViT-S/16 & 21.8M & - & - & - & 49.95 & 60.51 & 39.78 & 48.06 \\
Ours & ConvViT-S/16 & 21.9M & - & - & - & 52.01 & 60.77 & 45.84 & 54.11 \\
Ours & Swin-T/7 & 27.5M & - & - & - & 51.32 & 60.46 & 45.49 & 54.14 \\
\midrule
\multicolumn{10}{l}{Event-based self-supervised pre-training methods.} \\
\midrule
MEM~\cite{klenk2024masked} & ViT-B/16 & 85.6M & DSEC & 6000 & 70 & - & - & 44.62$^*$ & 51.39 \\
ECDP~\cite{yang2023event} & ViT-S/16 & 21.6M & N-ImageNet$^+$ & 300 & 267 & 54.66 & 66.08 & 47.91 & 56.50 \\
\midrule
MEM\#~\cite{klenk2024masked} & ViT-B/16 & 85.8M & N-ImageNet & 75 & 275 & 53.65 & 63.17 & 46.83 & 54.93 \\
ECDP\#~\cite{yang2023event} & ViT-S/16 & 21.6M & N-ImageNet$^+$ & 300 & 267 & 54.49 & 65.08 & 47.56 & 55.61 \\
\rowcolor{gray!20} Ours & ViT-S/16 & 21.8M & EF-ImageNet$^+$ & 710 & 80 & 56.06 & 65.27 & 47.24 & 55.53 \\
\midrule
ECDDP\#~\cite{yang2024event} & Swin-T/7 & 27.5M & E-TartanAir & 300 & 152 & 59.43 & 68.93 & 50.43 & 58.84 \\
\rowcolor{gray!20} Ours & ConvViT-S/16 & 21.9M & EF-ImageNet$^+$ & 710 & 88 & 60.53 & 70.09 & 53.22 & 61.84 \\
\rowcolor{gray!20} Ours & Swin-T/7 & 27.5M & EF-ImageNet$^+$ & 710 & 91 & 61.45 & 71.66 & 53.16 & 62.03 \\
\bottomrule
\end{tabularx}
\begin{tablenotes}[para,flushleft]
The indications of `+', `*', are the same as those in Table~\ref{mini_img}.
`\#': method fine-tuning using consistent base architectures and training method as ours.
\end{tablenotes}
\end{threeparttable}
\end{table*}

\begin{table*}[t]
\scriptsize
\centering
\caption{Full comparison of the performance on optical flow estimation tasks.}
\vspace{-3mm}
\label{flow_supp}
\setlength\tabcolsep{0.1cm}
\belowrulesep=0pt
\aboverulesep=0pt
\begin{threeparttable}
\begin{tabularx}{\linewidth}{>
{\centering\arraybackslash}p{3.0cm}|>
{\centering\arraybackslash}p{1.8cm}|>
{\centering\arraybackslash}p{1.8cm}>
{\centering\arraybackslash}p{1.0cm}|>
{\centering\arraybackslash}p{1.3cm}>
{\centering\arraybackslash}p{1.3cm}|>
{\centering\arraybackslash}p{1.38cm}>
{\centering\arraybackslash}p{1.38cm}|>
{\centering\arraybackslash}p{1.38cm}>
{\centering\arraybackslash}p{1.38cm}}
\toprule  
\multirow{2}{*}{Method} & \multirow{2}{*}{Backbone} & \multicolumn{2}{c|}{Pre-training} & \multicolumn{2}{c|}{indoor\_flying1} & \multicolumn{2}{c|}{indoor\_flying2} & \multicolumn{2}{c}{indoor\_flying3} \\
\cline{3-10}
 &  & Dataset & Time & AEE & Outlier(\%) & AEE & Outlier(\%) & AEE & Outlier(\%) \\
\midrule  
\multicolumn{10}{l}{Task-specific event-based methods.} \\
\midrule
EST~\cite{gehrig2019end} & - & - & - & 1.24 & 5.09 & 2.05 & 19.90 & 1.71 & 11.67 \\
DCEIFlow~\cite{wan2022learning} & - & - & - & \textbf{0.75}$^+$ & 0.60 & \textbf{1.39}$^+$ & 8.01 & \textbf{1.13}$^+$ & 5.29 \\
\midrule  
\multicolumn{10}{l}{Backbone training from scratch.} \\
\midrule
ECDP~\cite{yang2023event} & ViT-S/16 & - & - & 0.68 & 0.13 & 1.38 & 7.58 & 1.08 & 3.76 \\
Ours & ViT-S/16 & - & - & 0.734 & 1.087 & 1.630 & 15.890 & 1.292 & 6.563 \\
Ours & ConvViT-S/16 & - & - & 0.608 & 0.600 & 1.384 & 10.210 & 0.996 & 2.802 \\
Ours & Swin-T/7 & - & - & 0.491 & 0.002 & 1.110 & 7.187 & 0.804 & 1.865 \\
\midrule
\multicolumn{10}{l}{Event-based self-supervised pre-training methods.} \\
\midrule
ECDP~\cite{yang2023event} & ViT-S/16 & N-ImageNet$^+$ & 267 & 0.614 & 0.046 & 1.261 & 6.689 & 1.001 & 3.111 \\
\midrule
MEM\#~\cite{klenk2024masked} & ViTt-B/16 & N-ImageNet & 275 & 0.605 & 1.124 & 1.577 & 13.580 & 1.122 & 5.356 \\
ECDP\#~\cite{yang2023event} & ViT-S/16 & N-ImageNet$^+$ & 267 & 0.561 & 0.987 & 1.496 & 13.600 & 0.985 & 3.618 \\
\rowcolor{gray!20} Ours & ViT-S/16 & EF-ImageNet$^+$ & 80 & 0.457 & 0.015 & 1.076 & 4.855 & 0.768 & 0.812 \\
\midrule
ECDDP\#~\cite{yang2024event} & Swin-T/7 & E-TartanAir & 152 & 0.432 & 0.003 & 0.904 & 4.478 & 0.638 & 0.956 \\
\rowcolor{gray!20} Ours & ConvViT-S/16 & EF-ImageNet$^+$ & 88 & 0.434 & 0.004 & 1.035 & 5.041 & 0.778 & 0.305 \\
\rowcolor{gray!20} Ours & Swin-T/7 & EF-ImageNet$^+$ & 91 & 0.416 & 0.015 & 1.002 & 4.676 & 0.742 & 0.834 \\
\bottomrule
\end{tabularx}
\begin{tablenotes}[para,flushleft]
The indications of `+', `*', are the same as those in Table~\ref{mini_img}.
`\#' is the same as those in Table~\ref{sem_seg_supp}.
\end{tablenotes}
\end{threeparttable}
\end{table*}

\begin{table*}[t]
\scriptsize
\caption{Robustness study on object recognition tasks.}
\vspace{-3mm}
\label{robust_small}
\setlength\tabcolsep{0.1cm}
\belowrulesep=0pt
\aboverulesep=0pt
\begin{threeparttable}
\renewcommand{\arraystretch}{0.7}
\begin{tabularx}{\textwidth}{>
{\centering\arraybackslash}p{1.0cm}|>
{\centering\arraybackslash}p{0.8cm}|>
{\centering\arraybackslash}p{1.07cm}|>
{\centering\arraybackslash}p{0.9cm}|>
{\centering\arraybackslash}p{1.0cm}>
{\centering\arraybackslash}p{1.0cm}>
{\centering\arraybackslash}p{1.2cm}|>
{\centering\arraybackslash}p{2.0cm}|>
{\centering\arraybackslash}p{0.9cm}|>
{\centering\arraybackslash}p{1.0cm}>
{\centering\arraybackslash}p{1.0cm}>
{\centering\arraybackslash}p{1.2cm}|>
{\centering\arraybackslash}p{2.0cm}}
\toprule
\multirow{2}{*}{Method} & \multirow{2}{*}{Backbone} & Pre-training & \multicolumn{5}{c|}{N-Caltech101 Top-1 Acc(\%)} & \multicolumn{5}{c}{Mini-ES-ImageNet Top-1 Acc(\%)} \\
\cline{3-13}
 &  & Datasets & Org & Sparse & Noise & Sparse+Noise & [avg(Org-var) / Org]\%$\downarrow$ & Org & Sparse & Noise & Sparse+Noise & [avg(Org-var) / Org]\%$\downarrow$ \\
\midrule
MEM~\cite{klenk2024masked} & ViT-B & N-Img$^+$ & 86.55 & 85.36 & 85.71 & 84.66 & 1.510 & 54.75 & 54.02 & 54.26 & 53.79 & 1.327 \\
ECDP~\cite{yang2023event} & ViT-S & N-Img$^+$ & 85.64 & 83.28 & 84.26 & 84.32 & 1.969 & 66.02 & 64.27 & 64.63 & 63.49 & 2.863 \\
ECDP\&~\cite{yang2023event} & ViT-S & EF-Img$^+$ & - & - & - & - & - & 49.96 & 48.34 & 48.92 & 48.80 & 2.549 \\
ECDDP~\cite{yang2024event} & Swin-T & E-Tar & 82.78 & 79.63 & 81.80 & 80.27 & 2.964 & 62.77 & 61.44 & 61.20 & 61.08 & 2.437 \\
\rowcolor{gray!20}Ours & ViT-S & EF-Img$^+$ & 90.45 & 89.00 & 89.76 & 88.77 & 1.408 & 60.17 & 59.03 & 59.90 & 58.95 & 1.457 \\
\rowcolor{gray!20}Ours & CViT-S & EF-Img$^+$ & 92.36 & 91.15 & 92.31 & 91.67 & 0.769 & 66.86 & 66.25 & 65.89 & 65.65 & 1.391 \\
\rowcolor{gray!20}Ours & Swin-T & EF-Img$^+$ & 91.03 & 90.39 & 90.57 & 90.91 & 0.447 & 64.10 & 62.97 & 63.78 & 63.01 & 1.321 \\
\bottomrule
\end{tabularx}
\begin{tablenotes}[para,flushleft]
`Org': performance on original test set.
`Sparse', `noise' and `Sparse+Noise': Three manually created variant test sets.
`var': average performance on variant test sets (including `Sparse', `noise' and `Sparse+Noise').
\end{tablenotes}
\end{threeparttable}
\end{table*}

\begin{table*}[t]
\scriptsize
\centering
\caption{Hyparameters for pre-training.}
\vspace{-3mm}
\label{pr_params}
\setlength\tabcolsep{0.1cm}
\belowrulesep=0pt
\aboverulesep=0pt
\begin{threeparttable}
\begin{tabularx}{\linewidth}{>{\centering\arraybackslash}p{6.47cm}|>
{\centering\arraybackslash}p{3.5cm}>
{\centering\arraybackslash}p{3.5cm}>
{\centering\arraybackslash}p{3.5cm}}
\toprule
Hyperparameters & Masked Modeling & Feature Transition & Contrastive Learning \\
\midrule
batch size & 64 & 64 & 64 \\
warmup epochs & 20 & 2 & 20 \\
epochs & 500 & 10 & 200 \\
optimizer & AdamW & AdamW & AdamW \\
base learning rate & 1e-3 & 1e-4(5e-4) & 1e-3 \\
learning rate scheduler & cosine decay & cosine decay & cosine decay \\
learning rate layer decay & none & none & none \\
weight decay & 0.05 & 0.05 & 0.05 \\
drop path rate & none & none & none \\
grad clip & 5 & 5 & 5 \\
masking ratio & 0.50 & - & - \\
queue length & - & 1024 & 1024 \\
\bottomrule
\end{tabularx}
\begin{tablenotes}[para,flushleft]
Unbracketed cells: shared parameters for both ViT, ConvViT and Swin-Transformer.
Bracketed cells: parameters outside the brackets are for ViT and Swin-Transformer, inside are for ConvViT.
\end{tablenotes}
\end{threeparttable}
\end{table*}

\begin{table*}[t]
\scriptsize
\centering
\caption{Hyparameters for fine-tuning on object recognition tasks.}
\vspace{-3mm}
\label{obj_params}
\setlength\tabcolsep{0.1cm}
\belowrulesep=0pt
\aboverulesep=0pt
\begin{threeparttable}
\begin{tabularx}{\linewidth}{>
{\centering\arraybackslash}p{4.07cm}|>
{\centering\arraybackslash}p{2.5cm}>
{\centering\arraybackslash}p{2.5cm}>
{\centering\arraybackslash}p{2.5cm}|>
{\centering\arraybackslash}p{2.5cm}>
{\centering\arraybackslash}p{2.5cm}}
\toprule
Hyperparameters & N-Caltech101 & CIFAR10-DVS & N-Cars & N-ImageNet & ES-ImageNet \\
\midrule
batch size & 64 & 64 & 64 & 128 & 64 \\
warmup epochs & 20 & 20 & 20 & 20 & 20 \\
epochs & 300 & 300 & 300 & 100 & 100 \\
optimizer & AdamW & AdamW & AdamW & AdamW & AdamW \\
base learning rate & 2.5e-4 & 2.5e-4 & 2.5e-4 & 1e-4(5e-5) & 1e-3 \\
learning rate scheduler & cosine decay & cosine decay & cosine decay & cosine decay & cosine decay \\
learning rate layer decay & 0.75 & 0.75 & 0.75 & 0.75 & 0.75\\
weight decay & 0.05 & 0.05 & 0.05 & 0.05 & 0.05\\
drop path rate & 0.1 & 0.1 & 0.1 & 0.1 & 0.1 \\
grad clip & 5 & 5 & 5 & 5 & 5 \\
fixed events number & 30000 & 15000 & 3000 & 15000 & 20000 \\
\bottomrule
\end{tabularx}
\begin{tablenotes}[para,flushleft]
Unbracketed cells: shared parameters for both ViT, ConvViT and Swin-Transformer.
Bracketed cells: parameters outside the brackets are for ViT, inside are for ConvViT and Swin-Transformer.
\end{tablenotes}
\end{threeparttable}
\end{table*}

\begin{table*}[t]
\scriptsize
\centering
\caption{Hyparameters for fine-tuning on semantic segmentation, optical flow estimation and action recognition tasks.}
\vspace{-3mm}
\label{seg_flow_params}
\setlength\tabcolsep{0.1cm}
\belowrulesep=0pt
\aboverulesep=0pt
\begin{threeparttable}
\begin{tabularx}{\linewidth}{>
{\centering\arraybackslash}p{4.07cm}|>
{\centering\arraybackslash}p{2.5cm}>
{\centering\arraybackslash}p{2.5cm}|>
{\centering\arraybackslash}p{2.5cm}|>
{\centering\arraybackslash}p{2.5cm}>
{\centering\arraybackslash}p{2.5cm}}
\toprule
Hyperparameters & DDD17 & DSEC & MVSEC & DVS128 Gesture & UCF101-DVS \\
\midrule
batch size & 64 & 64 & 64 & 64 & 64 \\
warmup epochs & 20 & 20 & 20 & 20 & 20 \\
epochs & 100 & 100 & 150 & 300 & 300 \\
optimizer & AdamW & AdamW & AdamW & AdamW & AdamW \\
base learning rate & 1e-3 & 1e-3 & 1e-3 & 2.5e-4 & 2.5e-4 \\
learning rate scheduler & cosine decay & cosine decay & cosine decay & cosine decay & cosine decay \\
learning rate layer decay & 0.75 & 0.75 & 0.75 & 0.75 & 0.75 \\
weight decay & 0.05 & 0.05 & 0.05 & 0.05 & 0.05 \\
drop path rate & 0.1 & 0.1 & 0.1 & 0.1 & 0.1 \\
grad clip & 3 & 3 & 3 & 5 & 5 \\
fixed events number & 80000 & 200000 & - & 20000 & 40000 \\
dt & - & - & 1 & - & - \\
\bottomrule
\end{tabularx}
\begin{tablenotes}[para,flushleft]
\end{tablenotes}
\end{threeparttable}
\end{table*}

\end{document}